\documentclass[10pt,twocolumn,letterpaper]{article}

\usepackage{cvpr}              
\usepackage{stfloats}   

\usepackage[dvipsnames]{xcolor}

\usepackage{multirow}
\usepackage{bm}
\usepackage{pifont}
\usepackage{graphicx}
\usepackage{bbding}
\usepackage{tabularx}
\usepackage{booktabs}

\newcommand{\cparagraph}[1]{{\vspace{+1mm}\noindent\textbf{#1}\enspace}}

\definecolor{cvprblue}{rgb}{0.21,0.49,0.74}
\usepackage[pagebackref,breaklinks,colorlinks,citecolor=cvprblue]{hyperref}

\title{VoxHammer: Training-Free Precise and Coherent 3D Editing in Native 3D Space}

\author{
    {Lin Li\textsuperscript{2}\footnotemark[1] \quad
    Zehuan Huang\textsuperscript{1}\footnotemark[1]\footnotemark[2] \quad
    Haoran Feng\textsuperscript{3} \quad
    Gengxiong Zhuang\textsuperscript{1} \quad
    Rui Chen\textsuperscript{1}} \\
    \vspace{0.5em}{
    Chunchao Guo\textsuperscript{4} \quad
    Lu Sheng\textsuperscript{1 \Envelope}} \\
    \vspace{0.5em}{\textsuperscript{1}Beihang University \quad
    \textsuperscript{2}Renmin University of China \quad
    \textsuperscript{3}Tsinghua University \quad
    \textsuperscript{4}Tencent Hunyuan} \\
    {
    Project page: \url{https://huanngzh.github.io/VoxHammer-Page/}
    }
}

\begin{document}

\twocolumn[
    \maketitle
    \vspace{-2.8em}
    \begin{center}
\centering
\includegraphics[width=\textwidth]{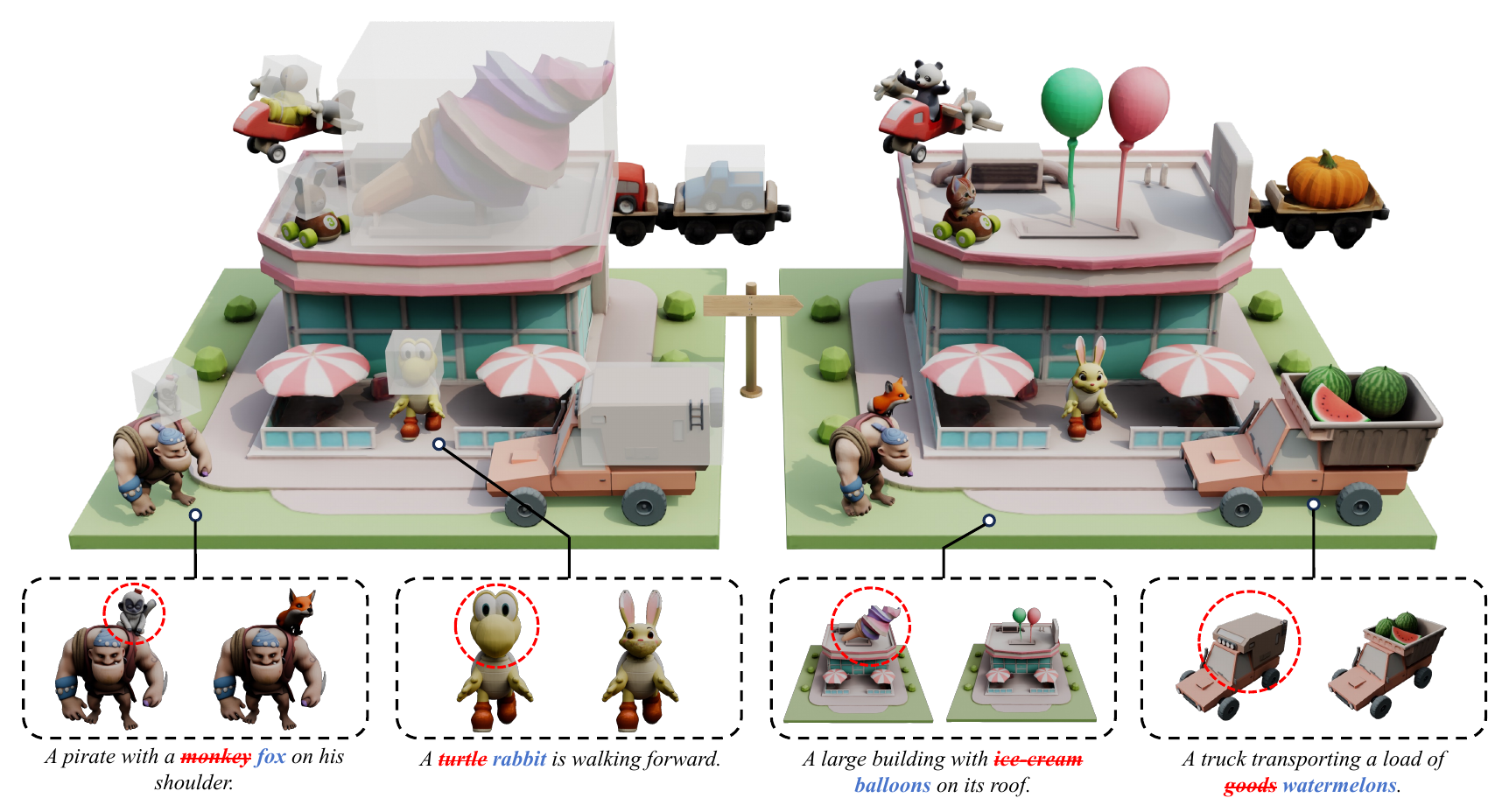}
\captionof{figure}{High-quality 3D assets edited by our method using text prompts. Our method uses a training-free approach to perform percise and coherent 3D local editing, transforming multiple 3D assets in the scene (left) into high-quality results (right). The bottom row shows a detailed comparison of each 3D asset before and after editing, as well as the conditioning texts.}
\label{fig:teaser}
\end{center}
    \bigbreak
]

\let\thefootnote\relax\footnotetext{
$^*$ Equal contribution \hspace{5pt}
$^\dagger$ Project lead \hspace{5pt} 
\ding{41} Corresponding author
}

\begin{abstract}
3D local editing of specified regions is crucial for game industry and robot interaction. Recent methods typically edit rendered multi-view images and then reconstruct 3D models, but they face challenges in precisely preserving unedited regions and overall coherence. Inspired by structured 3D generative models, we propose VoxHammer, a novel training-free approach that performs precise and coherent editing in 3D latent space. Given a 3D model, VoxHammer first predicts its inversion trajectory and obtains its inverted latents and key-value tokens at each timestep. Subsequently, in the denoising and editing phase, we replace the denoising features of preserved regions with the corresponding inverted latents and cached key-value tokens. By retaining these contextual features, this approach ensures consistent reconstruction of preserved areas and coherent integration of edited parts. To evaluate the consistency of preserved regions, we constructed Edit3D-Bench, a human-annotated dataset comprising hundreds of samples, each with carefully labeled 3D editing regions. Experiments demonstrate that VoxHammer significantly outperforms existing methods in terms of both 3D consistency of preserved regions and overall quality. Our method holds promise for synthesizing high-quality edited paired data, thereby laying the data foundation for in-context 3D generation.
\end{abstract}    
\section{Introduction}
\label{sec:intro}
In recent years, the rapid advancement of generative AI has greatly facilitated the creation of 3D assets~\cite{poole2022dreamfusiontextto3dusing2d, liu2024one2345, hong2023lrm, zhang2024clay, li2025triposg, trellis}, providing powerful production tools for industries such as gaming, robotics, and VR. Among these, 3D local editing~\cite{chen2023shapeditorinstructionguidedlatent3d, sella2023voxetextguidedvoxelediting, zhuang2024tipeditoraccurate3deditor,barda2024instant3ditmultiviewinpaintingfast,sella2025blendedpointcloud,meshpad} is a crucial task that enables partial modification of existing or AI-generated 3D assets while keeping other regions unchanged. It presents challenges in maintaining consistency in preserved regions and ensuring overall coherence in the edited model.

Existing 3D editing methods can be broadly categorized into two pipelines.
One approach~\cite{sella2023voxetextguidedvoxelediting,chen2023shapeditorinstructionguidedlatent3d,zhuang2024tipeditoraccurate3deditor,liu2024makeyour3dfastconsistentsubjectdriven,dong2024interactive3dcreatewantinteractive,dinh2025geometrystyle3dstylization} employs Score Distillation Sampling (SDS)~\cite{poole2022dreamfusiontextto3dusing2d} to optimize 3D representations so that it aligns with input prompts, but per-scene editing typically takes minutes or even hours.
Another approach~\cite{mvedit, baron2025editp233deditingpropagation,qi2024tailor3dcustomized3dassets,erkoç2024preditor3dfastprecise3d,gao20243dmesheditingusing,li2025cmdcontrollablemultiviewdiffusion} attempts to edit multi-view images~\cite{shi2023mvdream,huang2024mvadapter} rendered from the 3D model and then reconstruct the 3D model from the modified views.
These techniques achieve higher efficiency through a feed-forward process.
However, editing in 2D space instead of 3D space usually introduces position bias in the 3D reconstruction stage, making accurate local editing difficult.
In addition, inconsistencies among edited multi-view images~\cite{cao2024mvinpainterlearningmultiviewconsistent,barda2024instant3ditmultiviewinpaintingfast,erkoç2024preditor3dfastprecise3d} lead to artifacts in the edited 3D model, compromising quality and coherence~\cite{zheng2025pro3deditorprogressiveviewsperspective, li2025cmdcontrollablemultiviewdiffusion}.

Recently, advanced 3D generative models~\cite{zhang20233dshape2vecset,zhang2024clay,li2025triposg,wu2024direct3d,li2024craftsman,trellis,li2025sparc3d}, trained in native 3D space, can generate high-fidelity 3D content from text or image prompts.
These models exhibit significant advantages in 3D consistency and quality, inspiring us to edit 3D assets in native 3D space.
However, fine-tuning these models for editing is constrained by a critical data bottleneck: large-scale paired datasets for 3D local editing are exceptionally difficult to acquire.
Therefore, our research focuses on unleashing the potential of pretrained 3D generative models for precise and coherent 3D editing, eliminating the need for additional training.

We propose \textit{VoxHammer}, a training-free framework for precise and coherent 3D editing.
Our method is based on a pretrained structured 3D latent diffusion model~\cite{trellis}, and introduce a two-stage process: precise 3D inversion, and denoising and editing based on the inverted latents.
Given a 3D model (mesh, NeRF~\cite{mildenhall2021nerf}, or Gaussian Splat~\cite{kerbl20233dgs}), \textit{VoxHammer} first predicts its inversion trajectory of 3D diffusion process and caches its inverted latents and key-value tokens at each timestep.
We demonstrate that the inversion can reconstruct the given model's 3D geometry and texture with high precision.
In the subsequent denoising and editing phase, we denoise the edited region, and replace the denoising features of preserved regions with the corresponding inverted latents and cached key-value tokens.
By retaining these contextual features, our approach ensures consistent reconstruction of preserved areas and coherent integration of edited parts.
This is achieved without training the base model and only through feature replacement at inference, allowing high-quality 3D local editing at minimal cost.

The lack of labeled editing regions in existing datasets makes it challenging to objectively evaluate consistency in preserved areas. To address this, we constructed \textit{Edit3D-Bench}, a human-annotated dataset comprising hundreds of samples with carefully labeled 3D editing regions.
Quantitative and qualitative experiments on \textit{Edit3D-Bench} show that \textit{VoxHammer} significantly outperforms existing methods in terms of both editing accuracy and overall quality.
Our training-free method also holds promise for synthesizing high-quality edited paired data, thereby laying the data foundation for in-context 3D generation.
Our main contributions are summarized as follows:
\begin{itemize}
    \item We propose a training-free, native 3D local editing framework that leverages a pretrained 3D generative model for highly precise and coherent editing.
    \item We introduce precise 3D inversion and denoising-based editing using inverted latents, where we replace inverted latents and key-value tokens in the 3D latent space to ensure consistent reconstruction of preserved regions and coherent editing.
    \item We built \textit{Edit3D-Bench}, a benchmark to thoroughly validate the superiority of \textit{VoxHammer} in both editing accuracy and overall quality. Our method holds promise for synthesizing high-quality edited paired data, thereby laying the data foundation for in-context 3D generation.
\end{itemize}

\section{Related Work}
\label{sec:related_work}

\begin{figure*}
    \centering
    \includegraphics[width=\linewidth]{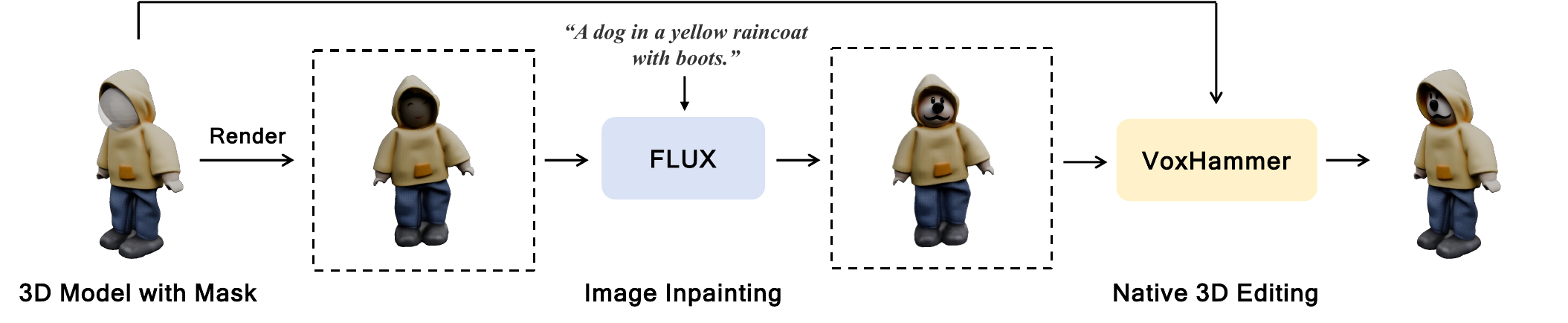}
    \caption{\textbf{Pipeline.} Given an input 3D model, a user-specified editing region, and a text prompt, the off-the-shelf models~\cite{flux2024,fluxfill} are used to inpaint the rendered view from the 3D model. Subsequently, our \textit{VoxHammer}, a training-free framework based on structured 3D diffusion models~\cite{trellis}, performs native 3D editing conditioned on the input 3D and the edited image.}
    \label{fig:pipeline}
\end{figure*}

\cparagraph{3D generative models.}
Recent advances in diffusion models~\cite{ho2020ddpm,song2020ddim} and the availability of high-quality 3D datasets~\cite{deitke2023objaverse,deitke2024objaversexl} have significantly accelerated the development of 3D generative models~\cite{liu2024one2345,liu2023syncdreamer,long2024wonder3d,hong2023lrm,tang2025lgm,huang2024epidiff,zhang2024clay,wu2024unique3d,li2024craftsman,wen2024ouroboros3d,xu2024instantmesh,voleti2025sv3d,wang2024crm,liu2024one2345++,wu2024direct3d,zhao2024michelangelo,roessle2024l3dg,wu2024blockfusion,meng2024lt3sd,liu2024part123,dong2025tela,chen2024meshxlneuralcoordinatefield,chen2024meshanythingartistcreatedmeshgeneration,wang2024llamameshunifying3dmesh,hao2024meshtronhighfidelityartistlike3d,he2024neurallightrigunlockingaccurate,gao2025meshartgeneratingarticulatedmeshes,zhao2025deepmeshautoregressiveartistmeshcreation,wei2025octgptoctreebasedmultiscaleautoregressive,li2025step1x3dhighfidelitycontrollablegeneration,ye2025shapellmomninativemultimodalllm}.
Some methods~\cite{liu2023syncdreamer,long2024wonder3d,tang2025lgm,wen2024ouroboros3d,xu2024instantmesh,wang2024crm,voleti2025sv3d,huang2024mvadapter,qu2025deocc1to33ddeocclusionsingle,huang2025stereogsmultiviewstereovision} generate 3D models by first synthesizing multi-view images and then reconstructing 3D from these views.
But inconsistent multi-view synthesis may lower the quality of the final 3D model.
A series of methods~\cite{zhang2024clay,li2024craftsman,wu2024direct3d,zhao2024michelangelo,li2025triposg,chen2025ultra3defficienthighfidelity3d,dong2025morecontextuallatents3d,zhao2025assemblerscalable3dassembly,tang2025efficientpartlevel3dobject,lin2025partcrafterstructured3dmesh,wu2025dipodualstateimagescontrolled,wu2025direct3ds2gigascale3dgeneration,li2025triposghighfidelity3dshape,trellis,li2025craftsman3dhighfidelitymeshgeneration} train native 3D generative models that comprise of a variational autoencoder~\cite{kingma2013vae} and a diffusion transformer (DiT)~\cite{peebles2023dit} for denoising in latent space.
These approaches unify 3D generation with high fidelity and consistency, laying the foundation for downstream inversion and editing.

\cparagraph{3D editing.}
Early 3D editing methods~\cite{chen2023shapeditorinstructionguidedlatent3d,sella2023voxetextguidedvoxelediting,zhuang2024tipeditoraccurate3deditor,liu2024makeyour3dfastconsistentsubjectdriven,dong2024interactive3dcreatewantinteractive,zhang2025scenelanguagerepresentingscenes,dinh2025geometrystyle3dstylization}  
employ Score Distillation Sampling (SDS)~\cite{poole2022dreamfusiontextto3dusing2d} to optimize 3D representation to align with the input prompts, but the per-scene editing requires minutes or even hours.
Subsequent works
~\cite{mvedit,qi2024tailor3dcustomized3dassets,cao2024mvinpainterlearningmultiviewconsistent,erkoç2024preditor3dfastprecise3d,gao20243dmesheditingusing,barda2024instant3ditmultiviewinpaintingfast,li2025cmdcontrollablemultiviewdiffusion,zheng2025pro3deditorprogressiveviewsperspective,baron2025editp233deditingpropagation} 
attempt to edit multi-view images rendered from the 3D model and reconstruct 3D from the modified views. 
But the lack of consistency across edited images frequently leads to degraded reconstruction quality.
In contrast, our method operates directly in the native 3D space, enabling precise and coherent editing that integrates seamlessly with the overall structure.

\cparagraph{Image generation and editing.}
Diffusion models~\cite{ho2020ddpm,song2020ddim,saharia2022imagen,ramesh2022dalle2,sd3,ldm,flux2024} synthesize images through a gradual denoising process starting from standard noise.
To enable image editing with pretrained image diffusion models, several methods~\cite{huberman2024ddpminversion,TamingRectifiedFlow,NullTextInversion,SemanticImageInversion,gal2022imageworthwordpersonalizing,zhu2025kvedittrainingfreeimageediting,dong2023prompttuninginversiontextdriven,feng2025personalizefreediffusiontransformer}
employ inversion techniques, which map real images into the diffusion model's denoising trajectory for precise controllable manipulation.
Inspired by these approaches, we explore inversion within native 3D generative models~\cite{trellis} and further propose a native 3D editing framework. 
Our method achieves training-free precise and coherent 3D editing through 3D inversion and novel contextual feature replacement.

\section{Methodology}

\begin{figure*}
    \centering
    \includegraphics[width=\linewidth]{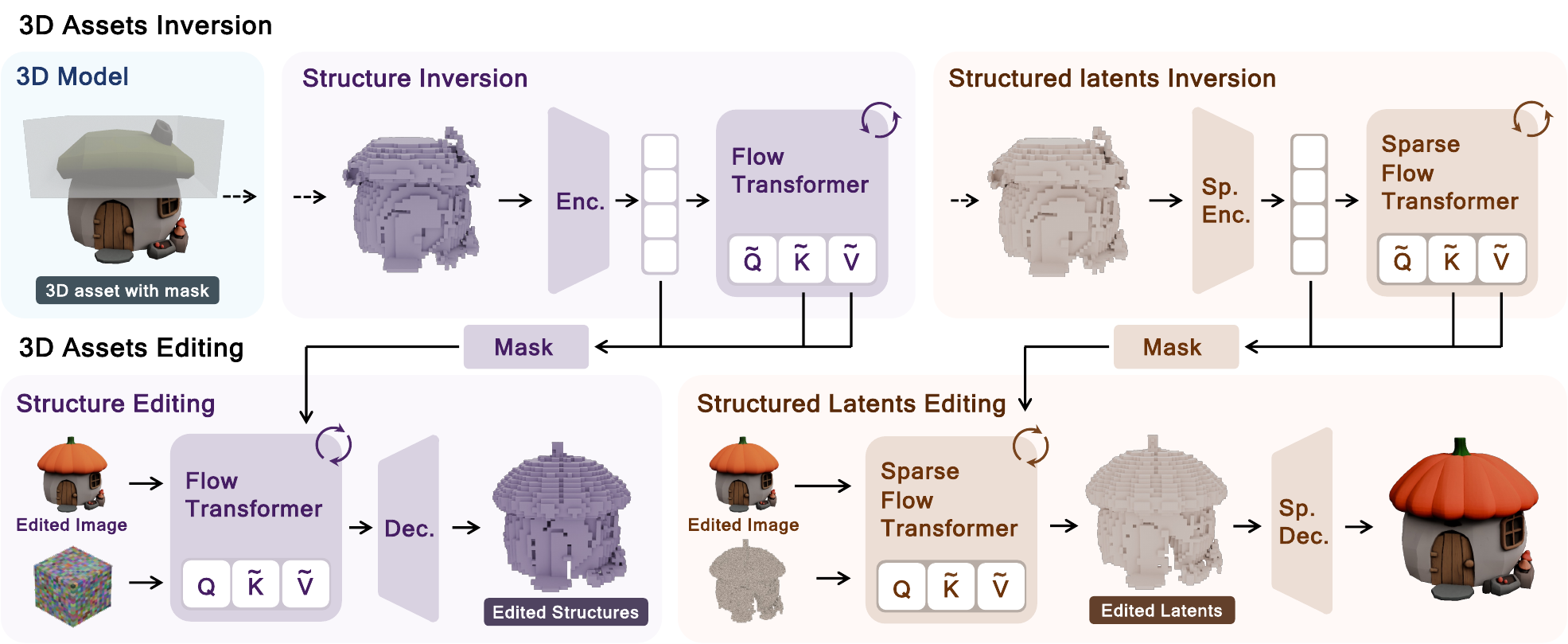}
    \caption{\textbf{Architecture of VoxHammer.} Our framework adopts TRELLIS~\cite{trellis} as the base model, which predicts sparse structures at the first structure (ST) stage and denoise fine-grained structured latents at the second sparse-latent (SLAT) stage. \textit{VoxHammer} performs inversion prediction in both the ST and SLAT stages, which map the textured 3D asset to its terminal noise, with latents and key/value tensors cached at each timestep. Subsequently, \textit{VoxHammer} denoises from the inverted noise, and replace the features of the preserved regions with the corresponding cached latents and key-value tokens, thereby achieving precise and coherent editing in native 3D space.}
    \label{fig:method}
\end{figure*}

We propose \textit{VoxHammer}, a training-free framework for 3D local editing, aimed at achieving precise and globally coherent modifications on 3D models.
As shown in \cref{fig:pipeline}, given an input 3D model (mesh, NeRF~\cite{mildenhall2021nerf}, or 3DGS~\cite{kerbl20233dgs}), an editing region and a text prompt, our framework first renders a view from the 3D model and utilizes advanced image diffusion models~\cite{flux2024,fluxfill} to produce an inpainted image. 
Subsequently, \textit{VoxHammer} performs native 3D editing conditioned on the input 3D model and the edited image.

The illustration of \textit{VoxHammer} is shown in \cref{fig:method}.
We first invert the 3D asset to noise and cache the inverted latents and key-value tokens at each timestep, which are then masked and used to guide the denoising process.
We first introduce our base model~\cite{trellis} and inversion design~\cite{TamingRectifiedFlow} in \cref{sec:preliminary}. 
Then, \cref{sec:3d_inversion} explores the 3D inversion and validates its consistency in geometry and texture reconstruction. 
Based on these findings, \cref{sec:3d_editing} presents our training-free 3D local editing method through inversion and re-editing.

\subsection{Preliminary}
\label{sec:preliminary}

\cparagraph{Structured 3D diffusion models.}
\textit{VoxHammer} is based on structured 3D latent diffusion models~\cite{trellis,wu2025amodal3ramodal3dreconstruction}, which are generative models that operate in a sparse voxel-based latent space for high-quality and scalable 3D generation. 
They represent a 3D asset (mesh, NeRF~\cite{mildenhall2021nerf}, or 3DGS~\cite{kerbl20233dgs}) as structured latents (SLAT), i.e., a set of local latent vectors $\{(z_i,p_i)\}_{i=1}^L$ anchored to active voxels $p_i$ that intersect the object surface, where each $z_i \in \mathbb{R}^C$ encodes fine-scale geometry and appearance.
During inference, the model first samples noise in the voxel-based latent space, followed by a two-stage denoising process.
In the first stage, referred to as the \textbf{structure (ST) stage}, the diffusion model predicts voxel occupancy over a $64^3$ grid to obtain sparse structures, where each location corresponds to a surface-intersecting voxel in the coarse space. 
In the second stage, called the \textbf{sparse-latent (SLAT) stage}, the structured latents are denoised to produce fine-grained geometry and texture, thereby enhancing the visual fidelity of the 3D output.

\cparagraph{Rectified Flow Inversion.}
Several methods~\cite{huberman2024ddpminversion,NullTextInversion,SemanticImageInversion,TamingRectifiedFlow} explore flow inversion for downstream editing tasks.
The challenges associated with achieving accurate inversion stem from the accumulation of numerical errors during ODE integration, which lead to noticeable deviations in the reconstructed samples. 
To address this, RF-Solver~\cite{TamingRectifiedFlow} introduces a training-free, plug-and-play sampler that improves inversion fidelity by analytically approximating the ODE solution using high-order Taylor expansion. 
This formulation significantly reduces integration errors and enables more faithful reconstructions.
Given a state $x^{\mathrm{ss}}_0$, it integrate the rectified-flow ODE from data to noise using a second-order Taylor-improved Euler scheme:
\begin{align}
x_{t-\Delta} &= x_t + \Delta\, f_\theta(x_t, t) + \frac{1}{2}\Delta^2 \,\partial_t f_\theta(x_t, t) \label{eq:taylor}\\
\partial_t f_\theta(x_t, t) &\approx \frac{f_\theta(x_{t-\Delta/2}, t-\Delta/2)-f_\theta(x_t, t)}{\Delta/2}
\end{align}
where $x_t$ represents the state of the data at the current time $t$, and $x_{t-\Delta}$ is the predicted state for the next denoising step, which is a time interval $\Delta$ away. The function $f_\theta(x_t, t)$ denotes the noise-prediction network, and $\partial_t f_\theta(x_t, t)$ is the second equation approximates using a finite difference scheme. The Taylor-improved update yields a local truncation error $\mathcal{O}(\Delta_t^3)$ and a global error $\mathcal{O}(\Delta_t^2)$, which it found crucial for high-fidelity reconstructions of fine structures.
Inspired by it, we explore inversion within native 3D generative models~\cite{trellis} and further propose our precise and coherent 3D editing framework through our novel contextual feature replacement.

\subsection{3D Inversion}
\label{sec:3d_inversion}

We introduce an inversion prediction strategy within the structured 3D generation pipeline~\cite{trellis} to map the textured 3D asset to its terminal noise.
The inversion proceeds in both the structure (ST) stage and the sparse-latent (SLAT) stage, with latents and key/value (K/V) tensors cached at each time step for later reuse during editing.

The inherent invertibility of the underlying Flux model enables this inversion strategy. In the forward pass, the pipeline generates assets by following a predefined discrete time schedule, denoted as $0=s_0<s_1<\dots<s_T=1$. To perform the inversion, we reverse the execution of this schedule by traversing the trajectory from timestep $s_T$ back to $s_0$. This reversed traversal deterministically traces the generation path backward, allowing us to map a final 3D asset to its corresponding source noise.

During the ST stage, the $K,V$ tensors from all attention layers are stored into a dictionary $\mathcal{KV}^{\mathrm{st}}$ indexed by latent time, block order, positional encoding, layer ID, and attention type.
In the SLAT stage, we first extract the preserved set $\Omega_{\mathrm{keep}}$ from the decoded output of the ST stage by removing the edit voxels, normalize the features, and run the same Taylor-improved inversion scheme. 
During this process, K/V tensors are cached into $\mathcal{KV}^{\mathrm{slat}}$. 
Throughout both stages, we apply classifier-free guidance (CFG)~\cite{ho2022cfg} only in a late-time interval $t \in [0.5, 1.0]$:
\begin{equation}
f_{\mathrm{cfg}} = (1+\omega)\, f_\theta(\mathrm{cond}) - \omega\,f_\theta(\mathrm{neg})
\end{equation}
and revert to $f_\theta(\mathrm{cond})$ otherwise, thereby stabilizing early inversion steps and improving fidelity. 
In practice, we keep $\omega$ fixed and activate guidance only in the late interval, which preserves the invertibility of early steps while providing sufficient semantic sharpness for the features in $\Omega_{\mathrm{keep}}$. 
During the next editing stage(\cref{sec:3d_editing}), the denoising features of the preserved regions are directly overwritten by the inverted source features, ensuring geometric and textural fidelity in unedited regions.

\begin{table*}
\small
\centering
\caption{\textbf{Quantitative comparison on our Edit3D-Bench}. We compute Chamfer Distance (CD.), masked PSNR, SSIM, LPIPS of unedited region to evaluate 3D consistency, FID, FVD to evaluate overall 3D quality, and DINO-I and CLIP-T to assess condition alignment.}
\label{tab:comparison} 
\setlength{\tabcolsep}{3.35mm} %
\resizebox{1.0\linewidth}{!}{
\begin{tabular}{l|cccc|cc|cc}
\toprule
\multirow{3}{*}[0.8ex]{Method} & \multicolumn{4}{c|}{Unedited Region Preservation} & \multicolumn{2}{c|}{Overall 3D Quality} &\multicolumn{2}{c}{Condition Alignment} \\
\cmidrule(lr){2-9} & CD. $\downarrow$ & PSNR (M) $\uparrow$ & SSIM (M) $\uparrow$ & LPIPS (M) $\downarrow$ & FID $\downarrow$ & FVD $\downarrow$  & DINO-I $\uparrow$ & CLIP-T $\uparrow$ \\
\midrule
Vox-E~\cite{sella2023voxetextguidedvoxelediting} & / & 13.84 & 0.827 & 0.316 & 87.41 & 3000.3 & 0.721 & 0.274 \\
MVEdit~\cite{mvedit} & 0.017 & 26.12 & 0.945 & 0.070 & 58.53 & 946.5 & 0.911 & 0.281 \\
Tailor3D~\cite{qi2024tailor3dcustomized3dassets} & 0.043 & 20.94 & 0.861 & 0.148 & 110.52 & 3812.1 & 0.704 & 0.258 \\
Instant3DiT~\cite{barda2024instant3ditmultiviewinpaintingfast} & 0.016 & 27.70 & 0.957 & 0.067 & 45.93 & 450.1 & 0.903 & 0.260 \\
TRELLIS~\cite{trellis} & 0.047 & 23.64 & 0.919 & 0.131 & 38.19 & 757.2 & 0.911 & 0.283 \\
\midrule
\textbf{Ours (full)} & \textbf{0.012} & \textbf{41.68} & \textbf{0.994} & \textbf{0.027} & \textbf{23.05} & \textbf{187.8} & \textbf{0.947} & \textbf{0.287} \\
\midrule
w/o Attn KV & 0.015 & 35.71 & 0.986 & 0.042 & 27.68 & 361.8 & 0.938 & 0.285 \\
w/ Noise Re-init & 0.014 & 36.31 & 0.989 & 0.038 & 25.71 & 259.6 & 0.945 & 0.287 \\
\bottomrule
\end{tabular}
}
\end{table*}

\subsection{3D Editing}
\label{sec:3d_editing}

Based on the inversion strategy, we further introduce a training-free local editing strategy, where the model denoises from the inverted noise, and performs latent replacement and key-value replacement on the unedited regions. 
Both operations are guided by 3D edit masks, to achieve precise preservation in unedited regions.

\cparagraph{Latent replacement.} 
In the structure (ST) stage, latent replacement is performed using a binary edit mask $M^{\mathrm{ss}} \in \{0,1\}^{H \times W \times D}$. 
At each denoising step $t$, the latent is blended with the inverted source latent $\hat{z}^{\mathrm{ss}}_t$:
\begin{equation}
z^{\mathrm{ss}}_{t} \leftarrow M^{\mathrm{ss}} \odot z^{\mathrm{ss}}_{t} + (1-M^{\mathrm{ss}}) \odot \hat{z}^{\mathrm{ss}}_{t}
\label{eq:ss_blend}
\end{equation}
To mitigate visible seams at mask boundaries, $M^{\mathrm{ss}}$ can be replaced with a soft mask $\widetilde{M}^{\mathrm{ss}}\in[0,1]^{H\times W\times D}$ obtained by dilation and Gaussian falloff.
In the sparse-latent (SLAT) stage, features at the unedited coordinate set $\Omega_{\mathrm{keep}}$ are replaced with the inverted source latent at each denoising step:
\begin{equation}
\forall\, \mathbf{u} \in \Omega_{\mathrm{keep}}: \quad z^{\mathrm{slat}}_{t}[\mathbf{u}] \leftarrow \hat{z}^{\mathrm{slat}}_{t}[\mathbf{u}]
\label{eq:slat_copy}
\end{equation}
Similar to the ST stage, coordinates near the boundaries can be weighted to ensure smooth transitions, effectively applying a soft-mask-like effect.

\cparagraph{Key-value replacement.}
Beyond latent replacement, feature-level consistency is enforced by our proposed key-value replacement in the attention mechanism.
In the ST stage, self--attention use binary masks $W^{\mathrm{self}}$ to indicate edited tokens. During editing, K/V tensors in unedited regions are replaced by their cached counterparts:
\begin{align}
K &\leftarrow W \odot K_{\mathrm{new}} + (1-W) \odot K_{\mathrm{cache}} \\
V &\leftarrow W \odot V_{\mathrm{new}} + (1-W) \odot V_{\mathrm{cache}}
\end{align}
Optional attention masks can be supplied to attention calculation to block mixing between edited and preserved tokens, which is especially helpful when the edited region is small but semantically strong.
Similar to the solution used in inversion, the model also adopts strategies of rescaled time scheduling and late-time CFG~\cite{ho2022cfg} in the denoise phase.
All the above modifications are implemented by dynamically adjusting forward functions at inference time, without retraining or weight updates.


\section{Experiments}
\label{sec:experiments}

\begin{figure*}
    \centering
    \includegraphics[width=\linewidth]{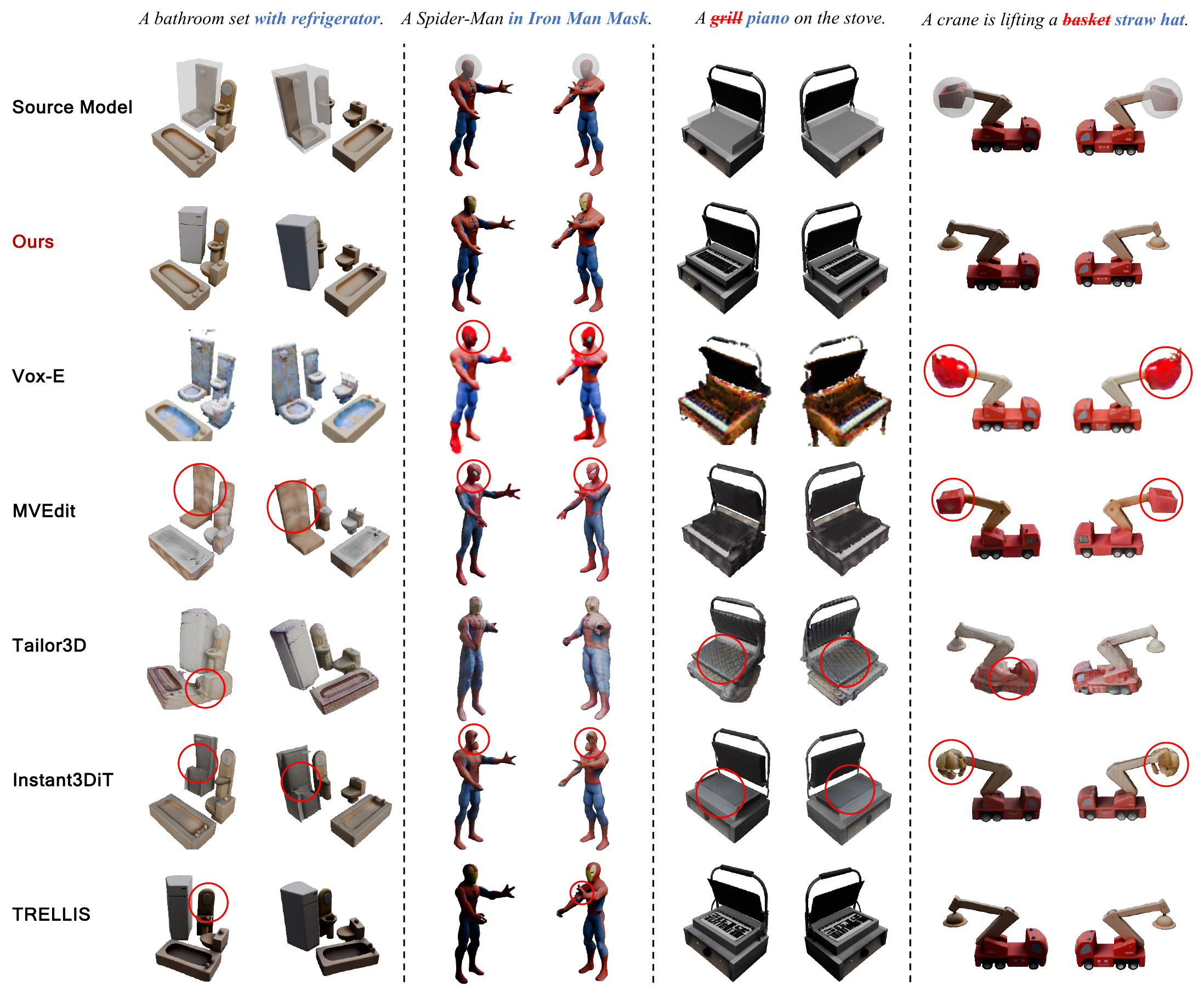}
    \caption{\textbf{Qualitative comparisons on Edit3D-Bench.} Our method achieves best performance on precision of editing and overall quality.}
    \label{fig:comparison}
\end{figure*}


\subsection{Setup}
\label{sec:setup} 

\cparagraph{Implementation details.} 
Our method is built on TRELLIS~\cite{trellis} and executed on a single NVIDIA A100 GPU. 
We set the sampling steps to 25 for both the inversion and denoising phases, and set the classifier-free guidance (CFG) scales of both stages to $5.0$ to balance reconstruction fidelity and edit creativity.



\cparagraph{Baselines.} 
We adopt five competitive 3D editing methods as baselines.
Vox-E~\cite{sella2023voxetextguidedvoxelediting} perform per-scene optimization on voxel representation with the guidance of image diffusion models.
MVEdit~\cite{mvedit}, Tailor3D~\cite{qi2024tailor3dcustomized3dassets} and Instant3dit~\cite{barda2024instant3ditmultiviewinpaintingfast} achieves customized 3D asset editing through multi-view editing.
TRELLIS~\cite{trellis} provides a native 3D editing method based on RePaint~\cite{repaint}, which guides the denoising process by sampling from the known regions.

\begin{figure*}
    \centering
    \includegraphics[width=\linewidth]{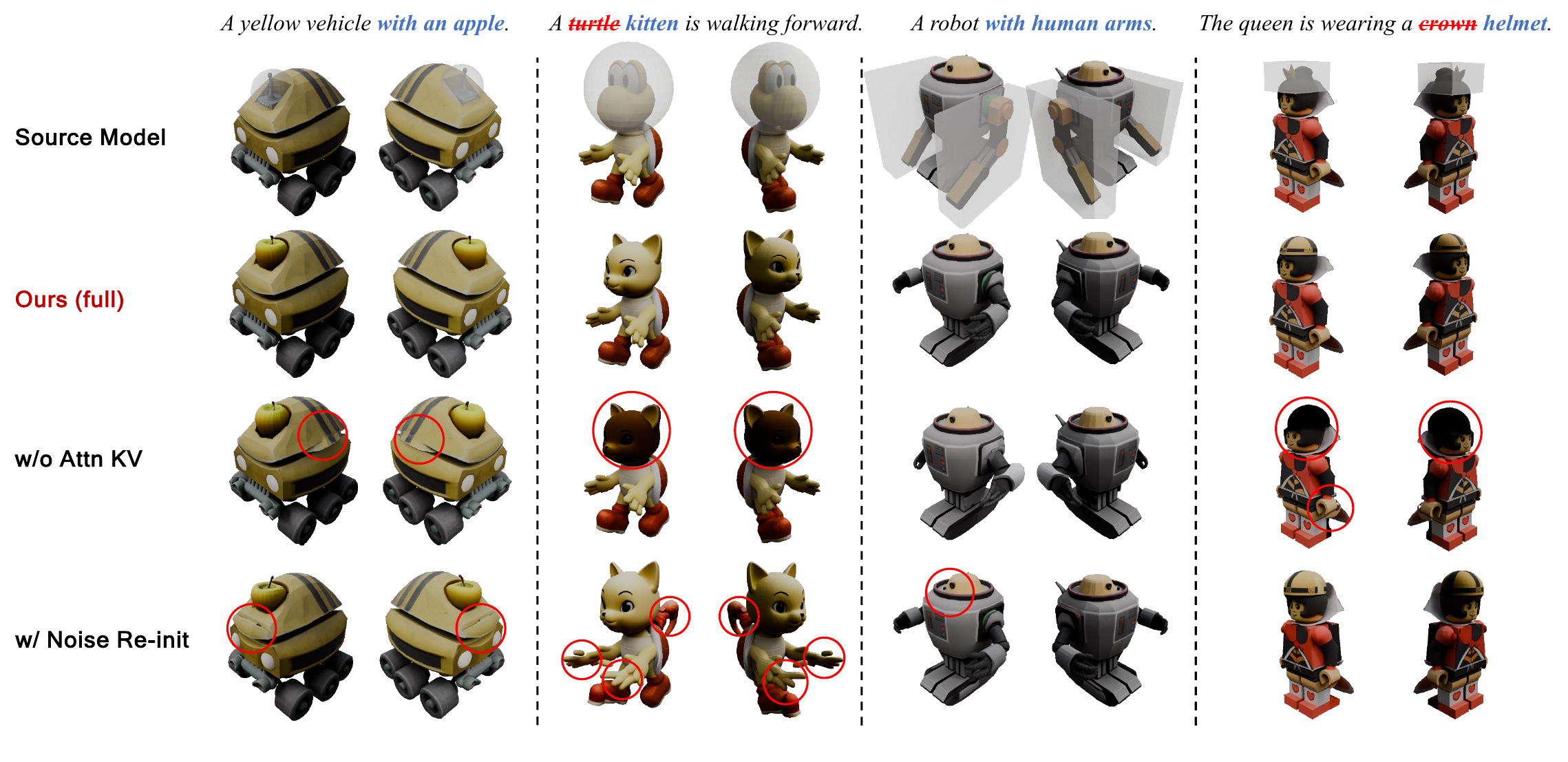}
    \caption{\textbf{Ablation studies.} Results demonstrate the effectiveness of key-value replacement in attention mechanism and latent replacement.}
    \label{fig:ablation}
\end{figure*}

\cparagraph{Evaluation dataset.} 
We conduct evaluation on \textit{Edit3D-Bench}, a new benchmark we curated for systematic 3D editing assessment.
It consists of 100 high-quality 3D models, with 50 carefully selected from Google Scanned Objects (GSO)~\cite{gso} and 50 from PartObjaverse-Tiny~\cite{partobjaverse_tiny}.
For each model, we provide 3 distinct editing prompts that cover a wide range of modifications. 
Each prompt is accompanied by a complete set of annotated 3D assets, including the 
2D renderings of the original object, 
a 2D mask of the edit region, 
a 2D edited image generated by FLUX.1 Fill~\cite{fluxfill}, illustrating the intended target edit, 
and a 3D mask specifying the precise editing region in 3D space. 
This well-structured dataset serves as a rigorous benchmark for assessing the accuracy, robustness, and fidelity of 3D editing methods.

\cparagraph{Evaluation metrics.} 
We use three aspects of metrics to comprehensively evaluate performance. 
First, for unedited region preservation, we assess the fidelity of preserved regions by computing Chamfer Distance (CD)~\cite{charmfer_distance} for geometry consistency, as well as masked PSNR, SSIM~\cite{ssim}, and LPIPS~\cite{lpips} on rendered multi-view images for texture.
Second, for overall 3D quality, we evaluate holistic performance by computing FID~\cite{fid} on rendered images and conducting a user study to capture human perceptual preferences.
Finally, for the alignment with the input prompts, we evaluate the alignment of the edited 3D assets with the edited image using DINO-I~\cite{dinov2}, and its alignment with the text prompt using CLIP-T~\cite{clip}.

\subsection{Main Results}
\label{sec:main_results}

\cparagraph{Quantitative comparison.}
As shown in ~\cref{tab:comparison}, our method significantly outperforms all baselines across nearly all metrics.
For unedited region preservation, our approach achieves the best scores in Chamfer Distance, PSNR, SSIM, and LPIPS, demonstrating its superior ability to maintain the original geometry and texture.
This is because our method operates directly in the native 3D space and leverages latents and key-value replacement to explicitly enforce consistency.
In contrast, multi-view based methods like MVEdit~\cite{mvedit} and Instant3DiT~\cite{barda2024instant3ditmultiviewinpaintingfast}, which rely on lifting multi-view edits to 3D space, usually introduce multi-view inconsistency and spatial bias, and struggle with maintaining 3D consistency.
TRELLIS~\cite{trellis} adapts Repaint~\cite{repaint} to achieve native 3D editing, but lacks inversion and key-value replacement to introduce the context of the reserved region, thus showing poor performance in 3D consistency.
For overall 3D quality and condition alignment, our method further demonstrates superiority by achieving the lowest FID and the highest DINO-I and CLIP-T scores. These results collectively indicate that our editing operations yield coherent and accurate outcomes in high-fidelity 3D models.

\begin{table}
\small
\centering
\caption{\textbf{User preference study.} Our method achieves consistently higher preference than Instant3DiT and TRELLIS in text alignment and overall 3D quality.}
\label{tab:user_study} 
\resizebox{1.0\linewidth}{!}{
\begin{tabular}{l|rr}
\toprule
Method & Text Alignment $\uparrow$ & Overall 3D Quality $\uparrow$ \\
\midrule
Instant3DiT~\cite{barda2024instant3ditmultiviewinpaintingfast} & 4.7\% &  1.6\% \\
TRELLIS~\cite{trellis} & 25.0\% &  17.2\% \\
\textbf{Ours} & \textbf{70.3\%} & \textbf{81.2\%} \\
\bottomrule
\end{tabular}
}
\end{table}

\begin{table}
\small
\centering
\caption{\textbf{Analysis of two-stage inversion.} We report Chamfer Distance (CD.) and PSNR, SSIM, LPIPS of rendered views to analyze the reconstruction consistency.}
\label{tab:ablation_inversion} 
\resizebox{1.0\linewidth}{!}{
\begin{tabular}{l|cccc}
\toprule
Inversion Stage & CD. $\downarrow$ & PSNR $\uparrow$ & SSIM $\uparrow$ & LPIPS $\downarrow$ \\
\midrule
ST stage & 0.0094 & 37.68 & 0.936 & 0.067 \\
ST + SLAT stage & \textbf{0.0055} & \textbf{39.70} & \textbf{0.987} & \textbf{0.012} \\
\bottomrule
\end{tabular}
}
\end{table}

\cparagraph{Qualitative comparison.} 
The qualitative results in ~\cref{fig:comparison} highlight the superiority of our method. 
Our approach consistently generates edits that are both locally accurate and geometrically coherent, while preserving unedited regions with high fidelity.
In contrast, the baselines exhibit various artifacts. 
Some methods suffer from poor reconstruction quality, leading to blurry outputs and distortions in preserved regions, as observed in Vox-E~\cite{sella2023voxetextguidedvoxelediting} and Tailor3D~\cite{qi2024tailor3dcustomized3dassets}.
Others are overly conservative, retaining most of the original content with only minimal edits, as is the case for MVEdit~\cite{mvedit}. 
Instant3DiT~\cite{barda2024instant3ditmultiviewinpaintingfast}, by contrast, often fails to generate results that align with text prompts, resulting in misplaced or misrepresented modifications.
In particular, the native TRELLIS~\cite{trellis} editing method, lacking inversion and KV-cache mechanisms, fails to effectively constrain the original 3D structure, often resulting in positional shifts and inconsistencies in the preserved regions.
Our method effectively avoids these issues, demonstrating the robustness of our native 3D, inversion-based editing framework.

\begin{figure}
    \centering
    \includegraphics[width=\linewidth]{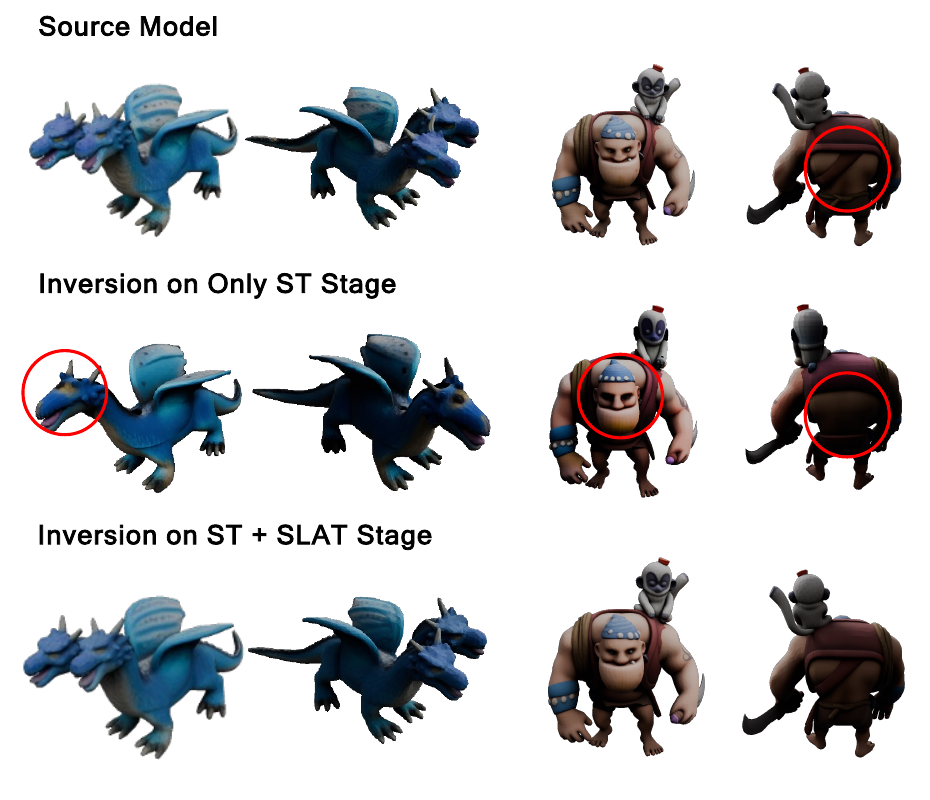}
    \caption{\textbf{The impact of inversion stages on reconstruction.} ST stage inversion lacks detailed consistency, while inversion on both stages achieves fine-grained geometry and texture reconstruction.}
    \label{fig:ablation_inversion}
\end{figure}

\begin{figure*}
    \centering
    \includegraphics[width=\linewidth]{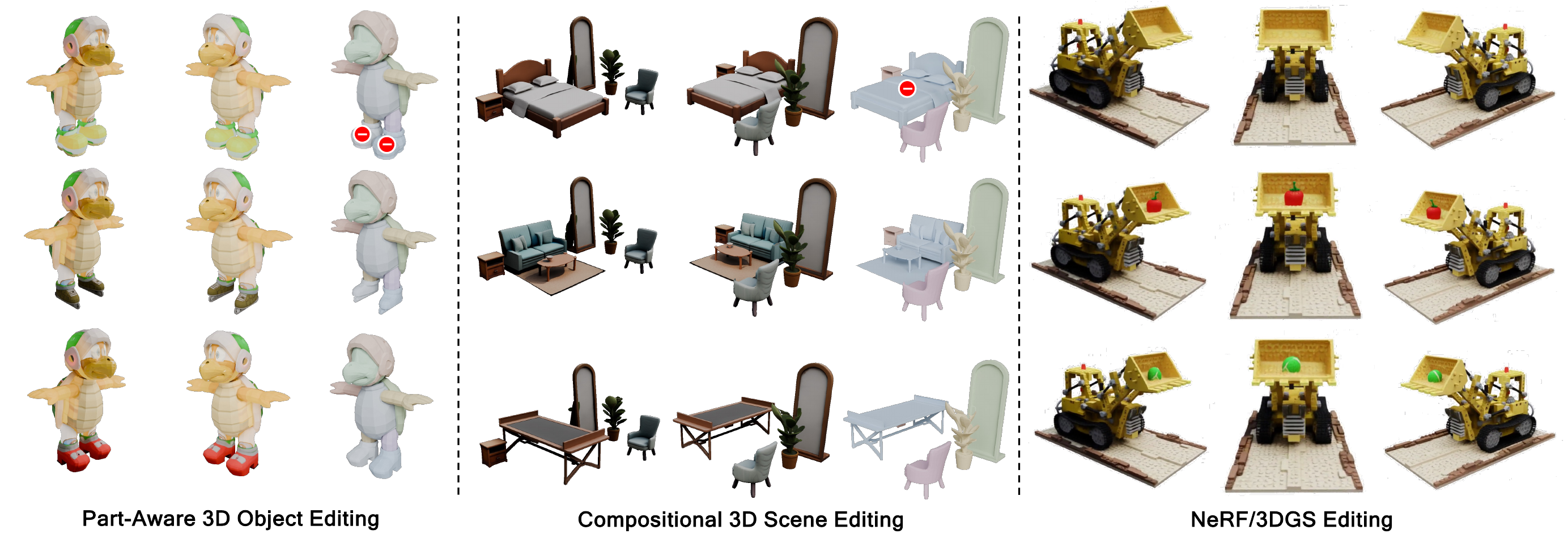}
    \caption{\textbf{More applications.} \textit{VoxHammer} easily generalizes to part-aware 3D object, scene, and NeRF~\cite{mildenhall2021nerf} or 3DGS~\cite{kerbl20233dgs} editing. We show the input models in the top row and the edited results in the bottom two rows.}
    \label{fig:applications}
\end{figure*}

\cparagraph{User study.} 
To evaluate the perceptual quality of our editing results, we conducted a user study with 30 participants. 
For each editing task, participants were presented with the input 3D models and the edit prompts, paired with the edited 3D assets by \textit{VoxHammer} alongside two strong baseline methods TRELLIS~\cite{trellis} and Instant3DiT~\cite{barda2024instant3ditmultiviewinpaintingfast}.
They were then asked to select the result that best aligned with the text prompt and exhibited the highest overall quality.
The study results demonstrate a clear human preference for our method, confirming its superior performance and robustness across multiple evaluation aspects.

\subsection{Ablation Study}
\label{sec:ablation} 

We conduct a series of ablation studies to analyze the reconstruction performance of two-stage inversion, and the contributions of the key components of our editing strategies, particularly the feature replacement strategy and the inversion-based initialization.

\cparagraph{Analysis of two-stage inversion.} 
We conducted analysis of doing inversion (1) \textit{only in ST stage}, and (2) \textit{in both ST and SLAT stage}, which is our full setting.
As shown in ~\cref{tab:ablation_inversion} and ~\cref{fig:ablation_inversion}, we found that the initial ST stage inversion provides a reasonable reconstruction of coarse geometry, but lacks detailed geometry and appearance consistency.
After incorporating the second SLAT stage, which handles high-resolution geometry and fine-grained texture details, the reconstruction quality improves significantly. 
It demonstrates that our two-stage inversion process faithfully restores the source 3D model with high fidelity, offering a robust basis for the subsequent editing stage.

\cparagraph{Assessment of editing strategies.}
We compare our full setting against four variants: 
(1) \textit{w/o Attn KV}, where we disable key-value replacement for attention mechanism; 
and (2) \textit{w/ Noise Re-init}, where we initiate the denoising process from randomly sampled Gaussian noise instead of the inverted noise of the source 3D asset.
The quantitative results are presented in ~\cref{tab:comparison}, and a qualitative comparison is shown in~\cref{fig:ablation}.
Missing key-value replacement leads to a significant degradation in preservation quality, as the edit concept leaks into unedited regions.
Re-initializing noise leads to a loss of positional information, causing unexpected alterations in preserved areas.
In contrast, our full setting effectively maintains the consistency of the preserved regions and achieves overall coherent editing, which validates that our inversion and key-value replacement is essential for achieving high-fidelity local 3D editing.

\subsection{More Applications}

\cparagraph{Part-aware object editing.}
\textit{VoxHammer} enables flexible editing of part-aware generated 3D assets~\cite{bang,yang2025omnipart,liu2024part123,partobjaverse_tiny,tang2025efficientpartlevel3dobject,lin2025partcrafterstructured3dmesh,chen2025autopartgen,dong2025copart,zhao2025assemblerscalable3dassembly}, where the pre-segmented structure offers 3D masks for us.
We report visualization results in ~\cref{fig:applications}.

\cparagraph{Compositional 3D scene editing.}
\textit{VoxHammer} further extends to compositional 3D scene editing~\cite{huang2025midi,yao2025cast}. 
As shown in ~\cref{fig:applications}, it supports fine-grained local modifications while preserving the integrity of the surrounding scene.

\cparagraph{NeRF or 3DGS editing.}
Benefiting from the versatility of the base model, \textit{VoxHammer} also generalizes NeRF~\cite{mildenhall2021nerf} or 3DGS~\cite{kerbl20233dgs} editing, as visualized in ~\cref{fig:applications}.

\section{Conclusion}
We presented \textit{VoxHammer}, a training-free framework for precise and coherent 3D local editing.
By leveraging accurate 3D inversion and feature replacement in the latent space of a pretrained structured 3D diffusion model, \textit{VoxHammer} preserves unedited regions with high fidelity while seamlessly integrating edits.
To enable objective evaluation, we introduced \textit{Edit3D-Bench}, a human-annotated benchmark for 3D local editing. Comprehensive experiments demonstrate that our method outperforms prior approaches in both consistency and quality. Beyond editing, \textit{VoxHammer} also enables the synthesis of paired data, laying a foundation for future in-context 3D generation.

{
    \small
    \bibliographystyle{ieeenat_fullname}
    \bibliography{main}

\begin{thebibliography}{106}
\providecommand{\natexlab}[1]{#1}
\providecommand{\url}[1]{\texttt{#1}}
\expandafter\ifx\csname urlstyle\endcsname\relax
  \providecommand{\doi}[1]{doi: #1}\else
  \providecommand{\doi}{doi: \begingroup \urlstyle{rm}\Url}\fi

\bibitem[Bar-On et~al.(2025)Bar-On, Cohen-Bar, and Cohen-Or]{baron2025editp233deditingpropagation}
Roi Bar-On, Dana Cohen-Bar, and Daniel Cohen-Or.
\newblock Editp23: 3d editing via propagation of image prompts to multi-view, 2025.

\bibitem[Barda et~al.(2024)Barda, Gadelha, Kim, Aigerman, Bermano, and Groueix]{barda2024instant3ditmultiviewinpaintingfast}
Amir Barda, Matheus Gadelha, Vladimir~G. Kim, Noam Aigerman, Amit~H. Bermano, and Thibault Groueix.
\newblock Instant3dit: Multiview inpainting for fast editing of 3d objects, 2024.

\bibitem[{Black Forest Labs}(2024)]{fluxfill}
{Black Forest Labs}.
\newblock {FLUX.1 Tools: Introducing Fill, Depth, Canny, and Redux}.
\newblock \url{https://bfl.ai/blog/24-11-21-tools}, 2024.
\newblock Accessed: 2025-08-15.

\bibitem[Cao et~al.(2024)Cao, Yu, Wang, Xue, and Fu]{cao2024mvinpainterlearningmultiviewconsistent}
Chenjie Cao, Chaohui Yu, Fan Wang, Xiangyang Xue, and Yanwei Fu.
\newblock Mvinpainter: Learning multi-view consistent inpainting to bridge 2d and 3d editing, 2024.

\bibitem[Chen et~al.(2024{\natexlab{a}})Chen, Shi, Liu, Shen, Gu, Wetzstein, Su, and Guibas]{mvedit}
Hansheng Chen, Ruoxi Shi, Yulin Liu, Bokui Shen, Jiayuan Gu, Gordon Wetzstein, Hao Su, and Leonidas Guibas.
\newblock Generic 3d diffusion adapter using controlled multi-view editing, 2024{\natexlab{a}}.

\bibitem[Chen et~al.(2023)Chen, Xie, Laina, and Vedaldi]{chen2023shapeditorinstructionguidedlatent3d}
Minghao Chen, Junyu Xie, Iro Laina, and Andrea Vedaldi.
\newblock Shap-editor: Instruction-guided latent 3d editing in seconds, 2023.

\bibitem[Chen et~al.(2025{\natexlab{a}})Chen, Wang, Shapovalov, Monnier, Jung, Wang, Ranjan, Laina, and Vedaldi]{chen2025autopartgen}
Minghao Chen, Jianyuan Wang, Roman Shapovalov, Tom Monnier, Hyunyoung Jung, Dilin Wang, Rakesh Ranjan, Iro Laina, and Andrea Vedaldi.
\newblock Autopartgen: Autogressive 3d part generation and discovery.
\newblock \emph{arXiv preprint arXiv:2507.13346}, 2025{\natexlab{a}}.

\bibitem[Chen et~al.(2024{\natexlab{b}})Chen, Chen, Pang, Zeng, Cheng, Fu, Yin, Wang, Wang, Zhang, Yu, Yu, Fu, and Chen]{chen2024meshxlneuralcoordinatefield}
Sijin Chen, Xin Chen, Anqi Pang, Xianfang Zeng, Wei Cheng, Yijun Fu, Fukun Yin, Yanru Wang, Zhibin Wang, Chi Zhang, Jingyi Yu, Gang Yu, Bin Fu, and Tao Chen.
\newblock Meshxl: Neural coordinate field for generative 3d foundation models, 2024{\natexlab{b}}.

\bibitem[Chen et~al.(2024{\natexlab{c}})Chen, He, Huang, Ye, Chen, Tang, Chen, Cai, Yang, Yu, Lin, and Zhang]{chen2024meshanythingartistcreatedmeshgeneration}
Yiwen Chen, Tong He, Di Huang, Weicai Ye, Sijin Chen, Jiaxiang Tang, Xin Chen, Zhongang Cai, Lei Yang, Gang Yu, Guosheng Lin, and Chi Zhang.
\newblock Meshanything: Artist-created mesh generation with autoregressive transformers, 2024{\natexlab{c}}.

\bibitem[Chen et~al.(2025{\natexlab{b}})Chen, Li, Wang, Zhang, Li, Zhang, and Lin]{chen2025ultra3defficienthighfidelity3d}
Yiwen Chen, Zhihao Li, Yikai Wang, Hu Zhang, Qin Li, Chi Zhang, and Guosheng Lin.
\newblock Ultra3d: Efficient and high-fidelity 3d generation with part attention, 2025{\natexlab{b}}.

\bibitem[Deitke et~al.(2023)Deitke, Schwenk, Salvador, Weihs, Michel, VanderBilt, Schmidt, Ehsani, Kembhavi, and Farhadi]{deitke2023objaverse}
Matt Deitke, Dustin Schwenk, Jordi Salvador, Luca Weihs, Oscar Michel, Eli VanderBilt, Ludwig Schmidt, Kiana Ehsani, Aniruddha Kembhavi, and Ali Farhadi.
\newblock Objaverse: A universe of annotated 3d objects.
\newblock In \emph{Proceedings of the IEEE/CVF Conference on Computer Vision and Pattern Recognition}, pages 13142--13153, 2023.

\bibitem[Deitke et~al.(2024)Deitke, Liu, Wallingford, Ngo, Michel, Kusupati, Fan, Laforte, Voleti, Gadre, et~al.]{deitke2024objaversexl}
Matt Deitke, Ruoshi Liu, Matthew Wallingford, Huong Ngo, Oscar Michel, Aditya Kusupati, Alan Fan, Christian Laforte, Vikram Voleti, Samir~Yitzhak Gadre, et~al.
\newblock Objaverse-xl: A universe of 10m+ 3d objects.
\newblock \emph{Advances in Neural Information Processing Systems}, 36, 2024.

\bibitem[Dinh et~al.(2025)Dinh, Lang, Kim, Stein, and Hanocka]{dinh2025geometrystyle3dstylization}
Nam~Anh Dinh, Itai Lang, Hyunwoo Kim, Oded Stein, and Rana Hanocka.
\newblock Geometry in style: 3d stylization via surface normal deformation, 2025.

\bibitem[Dong et~al.(2025{\natexlab{a}})Dong, Fang, Huang, Xu, Wang, Peng, and Dai]{dong2025tela}
Junting Dong, Qi Fang, Zehuan Huang, Xudong Xu, Jingbo Wang, Sida Peng, and Bo Dai.
\newblock Tela: Text to layer-wise 3d clothed human generation.
\newblock In \emph{European Conference on Computer Vision}, pages 19--36. Springer, 2025{\natexlab{a}}.

\bibitem[Dong et~al.(2024)Dong, Ding, Huang, Wang, Xue, and Xu]{dong2024interactive3dcreatewantinteractive}
Shaocong Dong, Lihe Ding, Zhanpeng Huang, Zibin Wang, Tianfan Xue, and Dan Xu.
\newblock Interactive3d: Create what you want by interactive 3d generation, 2024.

\bibitem[Dong et~al.(2025{\natexlab{b}})Dong, Ding, Chen, Li, Wang, Wang, Wang, Kim, Gao, Huang, Wang, Xue, and Xu]{dong2025morecontextuallatents3d}
Shaocong Dong, Lihe Ding, Xiao Chen, Yaokun Li, Yuxin Wang, Yucheng Wang, Qi Wang, Jaehyeok Kim, Chenjian Gao, Zhanpeng Huang, Zibin Wang, Tianfan Xue, and Dan Xu.
\newblock From one to more: Contextual part latents for 3d generation, 2025{\natexlab{b}}.

\bibitem[Dong et~al.(2025{\natexlab{c}})Dong, Ding, Chen, Li, Wang, Wang, Wang, Kim, Gao, Huang, et~al.]{dong2025copart}
Shaocong Dong, Lihe Ding, Xiao Chen, Yaokun Li, Yuxin Wang, Yucheng Wang, Qi Wang, Jaehyeok Kim, Chenjian Gao, Zhanpeng Huang, et~al.
\newblock From one to more: Contextual part latents for 3d generation.
\newblock \emph{arXiv preprint arXiv:2507.08772}, 2025{\natexlab{c}}.

\bibitem[Dong et~al.(2023)Dong, Xue, Duan, and Han]{dong2023prompttuninginversiontextdriven}
Wenkai Dong, Song Xue, Xiaoyue Duan, and Shumin Han.
\newblock Prompt tuning inversion for text-driven image editing using diffusion models, 2023.

\bibitem[Downs et~al.(2022)Downs, Francis, Koenig, Kinman, Hickman, Reymann, McHugh, and Vanhoucke]{gso}
Laura Downs, Anthony Francis, Nate Koenig, Brandon Kinman, Ryan Hickman, Krista Reymann, Thomas~B. McHugh, and Vincent Vanhoucke.
\newblock Google scanned objects: A high-quality dataset of 3d scanned household items, 2022.

\bibitem[Erkoç et~al.(2024)Erkoç, Gümeli, Wang, Nießner, Dai, Wonka, Lee, and Zhuang]{erkoç2024preditor3dfastprecise3d}
Ziya Erkoç, Can Gümeli, Chaoyang Wang, Matthias Nießner, Angela Dai, Peter Wonka, Hsin-Ying Lee, and Peiye Zhuang.
\newblock Preditor3d: Fast and precise 3d shape editing, 2024.

\bibitem[Esser et~al.(2024)Esser, Kulal, Blattmann, Entezari, Müller, Saini, Levi, Lorenz, Sauer, Boesel, Podell, Dockhorn, English, Lacey, Goodwin, Marek, and Rombach]{sd3}
Patrick Esser, Sumith Kulal, Andreas Blattmann, Rahim Entezari, Jonas Müller, Harry Saini, Yam Levi, Dominik Lorenz, Axel Sauer, Frederic Boesel, Dustin Podell, Tim Dockhorn, Zion English, Kyle Lacey, Alex Goodwin, Yannik Marek, and Robin Rombach.
\newblock Scaling rectified flow transformers for high-resolution image synthesis, 2024.

\bibitem[Fan et~al.(2016)Fan, Su, and Guibas]{charmfer_distance}
Haoqiang Fan, Hao Su, and Leonidas Guibas.
\newblock A point set generation network for 3d object reconstruction from a single image, 2016.

\bibitem[Feng et~al.(2025)Feng, Huang, Li, Lv, and Sheng]{feng2025personalizefreediffusiontransformer}
Haoran Feng, Zehuan Huang, Lin Li, Hairong Lv, and Lu Sheng.
\newblock Personalize anything for free with diffusion transformer, 2025.

\bibitem[Gal et~al.(2022)Gal, Alaluf, Atzmon, Patashnik, Bermano, Chechik, and Cohen-Or]{gal2022imageworthwordpersonalizing}
Rinon Gal, Yuval Alaluf, Yuval Atzmon, Or Patashnik, Amit~H. Bermano, Gal Chechik, and Daniel Cohen-Or.
\newblock An image is worth one word: Personalizing text-to-image generation using textual inversion, 2022.

\bibitem[Gao et~al.(2025)Gao, Siddiqui, Li, and Dai]{gao2025meshartgeneratingarticulatedmeshes}
Daoyi Gao, Yawar Siddiqui, Lei Li, and Angela Dai.
\newblock Meshart: Generating articulated meshes with structure-guided transformers, 2025.

\bibitem[Gao et~al.(2024)Gao, Wang, Fan, Bozic, Stuyck, Li, Dong, Ranjan, and Sarafianos]{gao20243dmesheditingusing}
Will Gao, Dilin Wang, Yuchen Fan, Aljaz Bozic, Tuur Stuyck, Zhengqin Li, Zhao Dong, Rakesh Ranjan, and Nikolaos Sarafianos.
\newblock 3d mesh editing using masked lrms, 2024.

\bibitem[Hao et~al.(2024)Hao, Romero, Lin, and Liu]{hao2024meshtronhighfidelityartistlike3d}
Zekun Hao, David~W. Romero, Tsung-Yi Lin, and Ming-Yu Liu.
\newblock Meshtron: High-fidelity, artist-like 3d mesh generation at scale, 2024.

\bibitem[He et~al.(2024)He, Wang, Huang, Pan, and Liu]{he2024neurallightrigunlockingaccurate}
Zexin He, Tengfei Wang, Xin Huang, Xingang Pan, and Ziwei Liu.
\newblock Neural lightrig: Unlocking accurate object normal and material estimation with multi-light diffusion, 2024.

\bibitem[Heusel et~al.(2018)Heusel, Ramsauer, Unterthiner, Nessler, and Hochreiter]{fid}
Martin Heusel, Hubert Ramsauer, Thomas Unterthiner, Bernhard Nessler, and Sepp Hochreiter.
\newblock Gans trained by a two time-scale update rule converge to a local nash equilibrium, 2018.

\bibitem[Ho and Salimans(2022)]{ho2022cfg}
Jonathan Ho and Tim Salimans.
\newblock Classifier-free diffusion guidance.
\newblock \emph{arXiv preprint arXiv:2207.12598}, 2022.

\bibitem[Ho et~al.(2020)Ho, Jain, and Abbeel]{ho2020ddpm}
Jonathan Ho, Ajay Jain, and Pieter Abbeel.
\newblock Denoising diffusion probabilistic models.
\newblock \emph{Advances in neural information processing systems}, 33:\penalty0 6840--6851, 2020.

\bibitem[Hong et~al.(2023)Hong, Zhang, Gu, Bi, Zhou, Liu, Liu, Sunkavalli, Bui, and Tan]{hong2023lrm}
Yicong Hong, Kai Zhang, Jiuxiang Gu, Sai Bi, Yang Zhou, Difan Liu, Feng Liu, Kalyan Sunkavalli, Trung Bui, and Hao Tan.
\newblock Lrm: Large reconstruction model for single image to 3d.
\newblock \emph{arXiv preprint arXiv:2311.04400}, 2023.

\bibitem[Huang et~al.(2025{\natexlab{a}})Huang, Cheung, Cong, See, and Wan]{huang2025stereogsmultiviewstereovision}
Xiufeng Huang, Ka~Chun Cheung, Runmin Cong, Simon See, and Renjie Wan.
\newblock Stereo-gs: Multi-view stereo vision model for generalizable 3d gaussian splatting reconstruction, 2025{\natexlab{a}}.

\bibitem[Huang et~al.(2024{\natexlab{a}})Huang, Guo, Wang, Yi, Ma, Cao, and Sheng]{huang2024mvadapter}
Zehuan Huang, Yuan-Chen Guo, Haoran Wang, Ran Yi, Lizhuang Ma, Yan-Pei Cao, and Lu Sheng.
\newblock Mv-adapter: Multi-view consistent image generation made easy.
\newblock \emph{arXiv preprint arXiv:2412.03632}, 2024{\natexlab{a}}.

\bibitem[Huang et~al.(2024{\natexlab{b}})Huang, Wen, Dong, Wang, Li, Chen, Cao, Liang, Qiao, Dai, et~al.]{huang2024epidiff}
Zehuan Huang, Hao Wen, Junting Dong, Yaohui Wang, Yangguang Li, Xinyuan Chen, Yan-Pei Cao, Ding Liang, Yu Qiao, Bo Dai, et~al.
\newblock Epidiff: Enhancing multi-view synthesis via localized epipolar-constrained diffusion.
\newblock In \emph{Proceedings of the IEEE/CVF Conference on Computer Vision and Pattern Recognition}, pages 9784--9794, 2024{\natexlab{b}}.

\bibitem[Huang et~al.(2025{\natexlab{b}})Huang, Guo, An, Yang, Li, Zou, Liang, Liu, Cao, and Sheng]{huang2025midi}
Zehuan Huang, Yuan-Chen Guo, Xingqiao An, Yunhan Yang, Yangguang Li, Zi-Xin Zou, Ding Liang, Xihui Liu, Yan-Pei Cao, and Lu Sheng.
\newblock Midi: Multi-instance diffusion for single image to 3d scene generation.
\newblock In \emph{Proceedings of the Computer Vision and Pattern Recognition Conference}, pages 23646--23657, 2025{\natexlab{b}}.

\bibitem[Huberman-Spiegelglas et~al.(2024)Huberman-Spiegelglas, Kulikov, and Michaeli]{huberman2024ddpminversion}
Inbar Huberman-Spiegelglas, Vladimir Kulikov, and Tomer Michaeli.
\newblock An edit friendly ddpm noise space: Inversion and manipulations.
\newblock In \emph{Proceedings of the IEEE/CVF Conference on Computer Vision and Pattern Recognition}, pages 12469--12478, 2024.

\bibitem[Kerbl et~al.(2023)Kerbl, Kopanas, Leimk{\"u}hler, and Drettakis]{kerbl20233dgs}
Bernhard Kerbl, Georgios Kopanas, Thomas Leimk{\"u}hler, and George Drettakis.
\newblock 3d gaussian splatting for real-time radiance field rendering.
\newblock \emph{ACM Trans. Graph.}, 42\penalty0 (4):\penalty0 139--1, 2023.

\bibitem[Kingma(2013)]{kingma2013vae}
Diederik~P Kingma.
\newblock Auto-encoding variational bayes.
\newblock \emph{arXiv preprint arXiv:1312.6114}, 2013.

\bibitem[Labs(2024)]{flux2024}
Black~Forest Labs.
\newblock Flux.
\newblock \url{https://github.com/black-forest-labs/flux}, 2024.

\bibitem[Li et~al.(2025{\natexlab{a}})Li, Erkoc, Li, Sirigatti, Rozov, Dai, and Nießner]{meshpad}
Haoxuan Li, Ziya Erkoc, Lei Li, Daniele Sirigatti, Vladyslav Rozov, Angela Dai, and Matthias Nießner.
\newblock Meshpad: Interactive sketch-conditioned artist-reminiscent mesh generation and editing, 2025{\natexlab{a}}.

\bibitem[Li et~al.(2025{\natexlab{b}})Li, Ma, Chen, Liu, Zhang, Xue, Luo, Sheffer, Wang, and Guo]{li2025cmdcontrollablemultiviewdiffusion}
Peng Li, Suizhi Ma, Jialiang Chen, Yuan Liu, Congyi Zhang, Wei Xue, Wenhan Luo, Alla Sheffer, Wenping Wang, and Yike Guo.
\newblock Cmd: Controllable multiview diffusion for 3d editing and progressive generation, 2025{\natexlab{b}}.

\bibitem[Li et~al.(2024)Li, Liu, Chen, Liang, Chen, Tan, and Long]{li2024craftsman}
Weiyu Li, Jiarui Liu, Rui Chen, Yixun Liang, Xuelin Chen, Ping Tan, and Xiaoxiao Long.
\newblock Craftsman: High-fidelity mesh generation with 3d native generation and interactive geometry refiner.
\newblock \emph{arXiv preprint arXiv:2405.14979}, 2024.

\bibitem[Li et~al.(2025{\natexlab{c}})Li, Liu, Yan, Chen, Liang, Chen, Tan, and Long]{li2025craftsman3dhighfidelitymeshgeneration}
Weiyu Li, Jiarui Liu, Hongyu Yan, Rui Chen, Yixun Liang, Xuelin Chen, Ping Tan, and Xiaoxiao Long.
\newblock Craftsman3d: High-fidelity mesh generation with 3d native generation and interactive geometry refiner, 2025{\natexlab{c}}.

\bibitem[Li et~al.(2025{\natexlab{d}})Li, Zhang, Sun, Qi, Li, Cheng, Cai, Wu, Liu, Wang, Chen, Tian, Pan, Li, Yu, Zhang, Jiang, and Tan]{li2025step1x3dhighfidelitycontrollablegeneration}
Weiyu Li, Xuanyang Zhang, Zheng Sun, Di Qi, Hao Li, Wei Cheng, Weiwei Cai, Shihao Wu, Jiarui Liu, Zihao Wang, Xiao Chen, Feipeng Tian, Jianxiong Pan, Zeming Li, Gang Yu, Xiangyu Zhang, Daxin Jiang, and Ping Tan.
\newblock Step1x-3d: Towards high-fidelity and controllable generation of textured 3d assets, 2025{\natexlab{d}}.

\bibitem[Li et~al.(2025{\natexlab{e}})Li, Zou, Liu, Wang, Liang, Yu, Liu, Guo, Liang, Ouyang, and Cao]{li2025triposghighfidelity3dshape}
Yangguang Li, Zi-Xin Zou, Zexiang Liu, Dehu Wang, Yuan Liang, Zhipeng Yu, Xingchao Liu, Yuan-Chen Guo, Ding Liang, Wanli Ouyang, and Yan-Pei Cao.
\newblock Triposg: High-fidelity 3d shape synthesis using large-scale rectified flow models, 2025{\natexlab{e}}.

\bibitem[Li et~al.(2025{\natexlab{f}})Li, Zou, Liu, Wang, Liang, Yu, Liu, Guo, Liang, Ouyang, et~al.]{li2025triposg}
Yangguang Li, Zi-Xin Zou, Zexiang Liu, Dehu Wang, Yuan Liang, Zhipeng Yu, Xingchao Liu, Yuan-Chen Guo, Ding Liang, Wanli Ouyang, et~al.
\newblock Triposg: High-fidelity 3d shape synthesis using large-scale rectified flow models.
\newblock \emph{arXiv preprint arXiv:2502.06608}, 2025{\natexlab{f}}.

\bibitem[Li et~al.(2025{\natexlab{g}})Li, Wang, Zheng, Luo, and Wen]{li2025sparc3d}
Zhihao Li, Yufei Wang, Heliang Zheng, Yihao Luo, and Bihan Wen.
\newblock Sparc3d: Sparse representation and construction for high-resolution 3d shapes modeling.
\newblock \emph{arXiv preprint arXiv:2505.14521}, 2025{\natexlab{g}}.

\bibitem[Lin et~al.(2025)Lin, Lin, Pan, Yan, Feng, Mu, and Fragkiadaki]{lin2025partcrafterstructured3dmesh}
Yuchen Lin, Chenguo Lin, Panwang Pan, Honglei Yan, Yiqiang Feng, Yadong Mu, and Katerina Fragkiadaki.
\newblock Partcrafter: Structured 3d mesh generation via compositional latent diffusion transformers, 2025.

\bibitem[Liu et~al.(2024{\natexlab{a}})Liu, Lin, Liu, Long, Dou, Guo, Luo, and Wang]{liu2024part123}
Anran Liu, Cheng Lin, Yuan Liu, Xiaoxiao Long, Zhiyang Dou, Hao-Xiang Guo, Ping Luo, and Wenping Wang.
\newblock Part123: part-aware 3d reconstruction from a single-view image.
\newblock In \emph{ACM SIGGRAPH 2024 Conference Papers}, pages 1--12, 2024{\natexlab{a}}.

\bibitem[Liu et~al.(2024{\natexlab{b}})Liu, Wang, Chen, Sun, and Duan]{liu2024makeyour3dfastconsistentsubjectdriven}
Fangfu Liu, Hanyang Wang, Weiliang Chen, Haowen Sun, and Yueqi Duan.
\newblock Make-your-3d: Fast and consistent subject-driven 3d content generation, 2024{\natexlab{b}}.

\bibitem[Liu et~al.(2024{\natexlab{c}})Liu, Shi, Chen, Zhang, Xu, Wei, Chen, Zeng, Gu, and Su]{liu2024one2345++}
Minghua Liu, Ruoxi Shi, Linghao Chen, Zhuoyang Zhang, Chao Xu, Xinyue Wei, Hansheng Chen, Chong Zeng, Jiayuan Gu, and Hao Su.
\newblock One-2-3-45++: Fast single image to 3d objects with consistent multi-view generation and 3d diffusion.
\newblock In \emph{Proceedings of the IEEE/CVF Conference on Computer Vision and Pattern Recognition}, pages 10072--10083, 2024{\natexlab{c}}.

\bibitem[Liu et~al.(2024{\natexlab{d}})Liu, Xu, Jin, Chen, Varma~T, Xu, and Su]{liu2024one2345}
Minghua Liu, Chao Xu, Haian Jin, Linghao Chen, Mukund Varma~T, Zexiang Xu, and Hao Su.
\newblock One-2-3-45: Any single image to 3d mesh in 45 seconds without per-shape optimization.
\newblock \emph{Advances in Neural Information Processing Systems}, 36, 2024{\natexlab{d}}.

\bibitem[Liu et~al.(2023)Liu, Lin, Zeng, Long, Liu, Komura, and Wang]{liu2023syncdreamer}
Yuan Liu, Cheng Lin, Zijiao Zeng, Xiaoxiao Long, Lingjie Liu, Taku Komura, and Wenping Wang.
\newblock Syncdreamer: Generating multiview-consistent images from a single-view image.
\newblock \emph{arXiv preprint arXiv:2309.03453}, 2023.

\bibitem[Long et~al.(2024)Long, Guo, Lin, Liu, Dou, Liu, Ma, Zhang, Habermann, Theobalt, et~al.]{long2024wonder3d}
Xiaoxiao Long, Yuan-Chen Guo, Cheng Lin, Yuan Liu, Zhiyang Dou, Lingjie Liu, Yuexin Ma, Song-Hai Zhang, Marc Habermann, Christian Theobalt, et~al.
\newblock Wonder3d: Single image to 3d using cross-domain diffusion.
\newblock In \emph{Proceedings of the IEEE/CVF Conference on Computer Vision and Pattern Recognition}, pages 9970--9980, 2024.

\bibitem[Lugmayr et~al.(2022)Lugmayr, Danelljan, Romero, Yu, Timofte, and Gool]{repaint}
Andreas Lugmayr, Martin Danelljan, Andres Romero, Fisher Yu, Radu Timofte, and Luc~Van Gool.
\newblock Repaint: Inpainting using denoising diffusion probabilistic models, 2022.

\bibitem[Meng et~al.(2024)Meng, Li, Nie{\ss}ner, and Dai]{meng2024lt3sd}
Quan Meng, Lei Li, Matthias Nie{\ss}ner, and Angela Dai.
\newblock Lt3sd: Latent trees for 3d scene diffusion.
\newblock \emph{arXiv preprint arXiv:2409.08215}, 2024.

\bibitem[Mildenhall et~al.(2021)Mildenhall, Srinivasan, Tancik, Barron, Ramamoorthi, and Ng]{mildenhall2021nerf}
Ben Mildenhall, Pratul~P Srinivasan, Matthew Tancik, Jonathan~T Barron, Ravi Ramamoorthi, and Ren Ng.
\newblock Nerf: Representing scenes as neural radiance fields for view synthesis.
\newblock \emph{Communications of the ACM}, 65\penalty0 (1):\penalty0 99--106, 2021.

\bibitem[Mokady et~al.(2023)Mokady, Hertz, Aberman, Pritch, and Cohen-Or]{NullTextInversion}
Ron Mokady, Amir Hertz, Kfir Aberman, Yael Pritch, and Daniel Cohen-Or.
\newblock Null-text inversion for editing real images using guided diffusion models.
\newblock In \emph{Proceedings of the IEEE/CVF Conference on Computer Vision and Pattern Recognition}, pages 6038--6047, 2023.

\bibitem[Oquab et~al.(2024)Oquab, Darcet, Moutakanni, Vo, Szafraniec, Khalidov, Fernandez, Haziza, Massa, El-Nouby, Assran, Ballas, Galuba, Howes, Huang, Li, Misra, Rabbat, Sharma, Synnaeve, Xu, Jegou, Mairal, Labatut, Joulin, and Bojanowski]{dinov2}
Maxime Oquab, Timothée Darcet, Théo Moutakanni, Huy Vo, Marc Szafraniec, Vasil Khalidov, Pierre Fernandez, Daniel Haziza, Francisco Massa, Alaaeldin El-Nouby, Mahmoud Assran, Nicolas Ballas, Wojciech Galuba, Russell Howes, Po-Yao Huang, Shang-Wen Li, Ishan Misra, Michael Rabbat, Vasu Sharma, Gabriel Synnaeve, Hu Xu, Hervé Jegou, Julien Mairal, Patrick Labatut, Armand Joulin, and Piotr Bojanowski.
\newblock Dinov2: Learning robust visual features without supervision, 2024.

\bibitem[Peebles and Xie(2023)]{peebles2023dit}
William Peebles and Saining Xie.
\newblock Scalable diffusion models with transformers.
\newblock In \emph{Proceedings of the IEEE/CVF International Conference on Computer Vision}, pages 4195--4205, 2023.

\bibitem[Poole et~al.(2022)Poole, Jain, Barron, and Mildenhall]{poole2022dreamfusiontextto3dusing2d}
Ben Poole, Ajay Jain, Jonathan~T. Barron, and Ben Mildenhall.
\newblock Dreamfusion: Text-to-3d using 2d diffusion, 2022.

\bibitem[Qi et~al.(2024)Qi, Yang, Zhang, Xing, Wu, Wu, Lin, Liu, Wang, and Zhao]{qi2024tailor3dcustomized3dassets}
Zhangyang Qi, Yunhan Yang, Mengchen Zhang, Long Xing, Xiaoyang Wu, Tong Wu, Dahua Lin, Xihui Liu, Jiaqi Wang, and Hengshuang Zhao.
\newblock Tailor3d: Customized 3d assets editing and generation with dual-side images, 2024.

\bibitem[Qu et~al.(2025)Qu, Dai, Li, Wang, Shen, Cao, and Ji]{qu2025deocc1to33ddeocclusionsingle}
Yansong Qu, Shaohui Dai, Xinyang Li, Yuze Wang, You Shen, Liujuan Cao, and Rongrong Ji.
\newblock Deocc-1-to-3: 3d de-occlusion from a single image via self-supervised multi-view diffusion, 2025.

\bibitem[Radford et~al.(2021)Radford, Kim, Hallacy, Ramesh, Goh, Agarwal, Sastry, Askell, Mishkin, Clark, Krueger, and Sutskever]{clip}
Alec Radford, Jong~Wook Kim, Chris Hallacy, Aditya Ramesh, Gabriel Goh, Sandhini Agarwal, Girish Sastry, Amanda Askell, Pamela Mishkin, Jack Clark, Gretchen Krueger, and Ilya Sutskever.
\newblock Learning transferable visual models from natural language supervision, 2021.

\bibitem[Ramesh et~al.(2022)Ramesh, Dhariwal, Nichol, Chu, and Chen]{ramesh2022dalle2}
Aditya Ramesh, Prafulla Dhariwal, Alex Nichol, Casey Chu, and Mark Chen.
\newblock Hierarchical text-conditional image generation with clip latents.
\newblock \emph{arXiv preprint arXiv:2204.06125}, 1\penalty0 (2):\penalty0 3, 2022.

\bibitem[Roessle et~al.(2024)Roessle, M{\"u}ller, Porzi, Bul{\o}, Kontschieder, Dai, and Nie{\ss}ner]{roessle2024l3dg}
Barbara Roessle, Norman M{\"u}ller, Lorenzo Porzi, Samuel~Rota Bul{\o}, Peter Kontschieder, Angela Dai, and Matthias Nie{\ss}ner.
\newblock L3dg: Latent 3d gaussian diffusion.
\newblock \emph{arXiv preprint arXiv:2410.13530}, 2024.

\bibitem[Rombach et~al.(2022)Rombach, Blattmann, Lorenz, Esser, and Ommer]{ldm}
Robin Rombach, Andreas Blattmann, Dominik Lorenz, Patrick Esser, and Björn Ommer.
\newblock High-resolution image synthesis with latent diffusion models, 2022.

\bibitem[Rout et~al.(2024)Rout, Chen, Ruiz, Caramanis, Shakkottai, and Chu]{SemanticImageInversion}
Litu Rout, Yujia Chen, Nataniel Ruiz, Constantine Caramanis, Sanjay Shakkottai, and Wen-Sheng Chu.
\newblock Semantic image inversion and editing using rectified stochastic differential equations.
\newblock \emph{arXiv preprint arXiv:2410.10792}, 2024.

\bibitem[Saharia et~al.(2022)Saharia, Chan, Saxena, Li, Whang, Denton, Ghasemipour, Gontijo~Lopes, Karagol~Ayan, Salimans, et~al.]{saharia2022imagen}
Chitwan Saharia, William Chan, Saurabh Saxena, Lala Li, Jay Whang, Emily~L Denton, Kamyar Ghasemipour, Raphael Gontijo~Lopes, Burcu Karagol~Ayan, Tim Salimans, et~al.
\newblock Photorealistic text-to-image diffusion models with deep language understanding.
\newblock \emph{Advances in neural information processing systems}, 35:\penalty0 36479--36494, 2022.

\bibitem[Sella et~al.(2023)Sella, Fiebelman, Hedman, and Averbuch-Elor]{sella2023voxetextguidedvoxelediting}
Etai Sella, Gal Fiebelman, Peter Hedman, and Hadar Averbuch-Elor.
\newblock Vox-e: Text-guided voxel editing of 3d objects, 2023.

\bibitem[Sella et~al.(2025)Sella, Atia, Mokady, and Averbuch-Elor]{sella2025blendedpointcloud}
Etai Sella, Noam Atia, Ron Mokady, and Hadar Averbuch-Elor.
\newblock Blended point cloud diffusion for localized text-guided shape editing.
\newblock \emph{arXiv preprint arXiv:2507.15399}, 2025.

\bibitem[Shi et~al.(2023)Shi, Wang, Ye, Long, Li, and Yang]{shi2023mvdream}
Yichun Shi, Peng Wang, Jianglong Ye, Mai Long, Kejie Li, and Xiao Yang.
\newblock Mvdream: Multi-view diffusion for 3d generation.
\newblock \emph{arXiv preprint arXiv:2308.16512}, 2023.

\bibitem[Song et~al.(2020)Song, Meng, and Ermon]{song2020ddim}
Jiaming Song, Chenlin Meng, and Stefano Ermon.
\newblock Denoising diffusion implicit models.
\newblock \emph{arXiv preprint arXiv:2010.02502}, 2020.

\bibitem[Tang et~al.(2025{\natexlab{a}})Tang, Chen, Chen, Wang, Zeng, and Liu]{tang2025lgm}
Jiaxiang Tang, Zhaoxi Chen, Xiaokang Chen, Tengfei Wang, Gang Zeng, and Ziwei Liu.
\newblock Lgm: Large multi-view gaussian model for high-resolution 3d content creation.
\newblock In \emph{European Conference on Computer Vision}, pages 1--18. Springer, 2025{\natexlab{a}}.

\bibitem[Tang et~al.(2025{\natexlab{b}})Tang, Lu, Li, Hao, Li, Wei, Song, Zeng, Liu, and Lin]{tang2025efficientpartlevel3dobject}
Jiaxiang Tang, Ruijie Lu, Zhaoshuo Li, Zekun Hao, Xuan Li, Fangyin Wei, Shuran Song, Gang Zeng, Ming-Yu Liu, and Tsung-Yi Lin.
\newblock Efficient part-level 3d object generation via dual volume packing, 2025{\natexlab{b}}.

\bibitem[Voleti et~al.(2025)Voleti, Yao, Boss, Letts, Pankratz, Tochilkin, Laforte, Rombach, and Jampani]{voleti2025sv3d}
Vikram Voleti, Chun-Han Yao, Mark Boss, Adam Letts, David Pankratz, Dmitry Tochilkin, Christian Laforte, Robin Rombach, and Varun Jampani.
\newblock Sv3d: Novel multi-view synthesis and 3d generation from a single image using latent video diffusion.
\newblock In \emph{European Conference on Computer Vision}, pages 439--457. Springer, 2025.

\bibitem[Wang et~al.(2024{\natexlab{a}})Wang, Pu, Qi, Guo, Ma, Huang, Chen, Li, and Shan]{TamingRectifiedFlow}
Jiangshan Wang, Junfu Pu, Zhongang Qi, Jiayi Guo, Yue Ma, Nisha Huang, Yuxin Chen, Xiu Li, and Ying Shan.
\newblock Taming rectified flow for inversion and editing.
\newblock \emph{arXiv preprint arXiv:2411.04746}, 2024{\natexlab{a}}.

\bibitem[Wang et~al.(2004)Wang, Bovik, Sheikh, and Simoncelli]{ssim}
Zhou Wang, A.C. Bovik, H.R. Sheikh, and E.P. Simoncelli.
\newblock Image quality assessment: from error visibility to structural similarity.
\newblock \emph{IEEE Transactions on Image Processing}, 13\penalty0 (4):\penalty0 600--612, 2004.

\bibitem[Wang et~al.(2024{\natexlab{b}})Wang, Lorraine, Wang, Su, Zhu, Fidler, and Zeng]{wang2024llamameshunifying3dmesh}
Zhengyi Wang, Jonathan Lorraine, Yikai Wang, Hang Su, Jun Zhu, Sanja Fidler, and Xiaohui Zeng.
\newblock Llama-mesh: Unifying 3d mesh generation with language models, 2024{\natexlab{b}}.

\bibitem[Wang et~al.(2024{\natexlab{c}})Wang, Wang, Chen, Xiang, Chen, Yu, Li, Su, and Zhu]{wang2024crm}
Zhengyi Wang, Yikai Wang, Yifei Chen, Chendong Xiang, Shuo Chen, Dajiang Yu, Chongxuan Li, Hang Su, and Jun Zhu.
\newblock Crm: Single image to 3d textured mesh with convolutional reconstruction model.
\newblock \emph{arXiv preprint arXiv:2403.05034}, 2024{\natexlab{c}}.

\bibitem[Wei et~al.(2025)Wei, Wang, Zhou, Chen, and Wang]{wei2025octgptoctreebasedmultiscaleautoregressive}
Si-Tong Wei, Rui-Huan Wang, Chuan-Zhi Zhou, Baoquan Chen, and Peng-Shuai Wang.
\newblock Octgpt: Octree-based multiscale autoregressive models for 3d shape generation, 2025.

\bibitem[Wen et~al.(2024)Wen, Huang, Wang, Chen, Qiao, and Sheng]{wen2024ouroboros3d}
Hao Wen, Zehuan Huang, Yaohui Wang, Xinyuan Chen, Yu Qiao, and Lu Sheng.
\newblock Ouroboros3d: Image-to-3d generation via 3d-aware recursive diffusion.
\newblock \emph{arXiv preprint arXiv:2406.03184}, 2024.

\bibitem[Wu et~al.(2024{\natexlab{a}})Wu, Liu, Cai, Yan, Wang, Hu, Duan, and Ma]{wu2024unique3d}
Kailu Wu, Fangfu Liu, Zhihan Cai, Runjie Yan, Hanyang Wang, Yating Hu, Yueqi Duan, and Kaisheng Ma.
\newblock Unique3d: High-quality and efficient 3d mesh generation from a single image.
\newblock \emph{arXiv preprint arXiv:2405.20343}, 2024{\natexlab{a}}.

\bibitem[Wu et~al.(2025{\natexlab{a}})Wu, Wang, Liu, Guo, Qiu, Li, Huang, Su, and Cheng]{wu2025dipodualstateimagescontrolled}
Ruiqi Wu, Xinjie Wang, Liu Liu, Chunle Guo, Jiaxiong Qiu, Chongyi Li, Lichao Huang, Zhizhong Su, and Ming-Ming Cheng.
\newblock Dipo: Dual-state images controlled articulated object generation powered by diverse data, 2025{\natexlab{a}}.

\bibitem[Wu et~al.(2024{\natexlab{b}})Wu, Lin, Zhang, Zeng, Xu, Torr, Cao, and Yao]{wu2024direct3d}
Shuang Wu, Youtian Lin, Feihu Zhang, Yifei Zeng, Jingxi Xu, Philip Torr, Xun Cao, and Yao Yao.
\newblock Direct3d: Scalable image-to-3d generation via 3d latent diffusion transformer.
\newblock \emph{arXiv preprint arXiv:2405.14832}, 2024{\natexlab{b}}.

\bibitem[Wu et~al.(2025{\natexlab{b}})Wu, Lin, Zhang, Zeng, Yang, Bao, Qian, Zhu, Cao, Torr, and Yao]{wu2025direct3ds2gigascale3dgeneration}
Shuang Wu, Youtian Lin, Feihu Zhang, Yifei Zeng, Yikang Yang, Yajie Bao, Jiachen Qian, Siyu Zhu, Xun Cao, Philip Torr, and Yao Yao.
\newblock Direct3d-s2: Gigascale 3d generation made easy with spatial sparse attention, 2025{\natexlab{b}}.

\bibitem[Wu et~al.(2025{\natexlab{c}})Wu, Zheng, Guan, Vedaldi, and Cham]{wu2025amodal3ramodal3dreconstruction}
Tianhao Wu, Chuanxia Zheng, Frank Guan, Andrea Vedaldi, and Tat-Jen Cham.
\newblock Amodal3r: Amodal 3d reconstruction from occluded 2d images, 2025{\natexlab{c}}.

\bibitem[Wu et~al.(2024{\natexlab{c}})Wu, Li, Yan, Shang, Sun, Wang, Cui, Liu, Sato, Li, et~al.]{wu2024blockfusion}
Zhennan Wu, Yang Li, Han Yan, Taizhang Shang, Weixuan Sun, Senbo Wang, Ruikai Cui, Weizhe Liu, Hiroyuki Sato, Hongdong Li, et~al.
\newblock Blockfusion: Expandable 3d scene generation using latent tri-plane extrapolation.
\newblock \emph{ACM Transactions on Graphics (TOG)}, 43\penalty0 (4):\penalty0 1--17, 2024{\natexlab{c}}.

\bibitem[Xiang et~al.(2025)Xiang, Lv, Xu, Deng, Wang, Zhang, Chen, Tong, and Yang]{trellis}
Jianfeng Xiang, Zelong Lv, Sicheng Xu, Yu Deng, Ruicheng Wang, Bowen Zhang, Dong Chen, Xin Tong, and Jiaolong Yang.
\newblock Structured 3d latents for scalable and versatile 3d generation, 2025.

\bibitem[Xu et~al.(2024)Xu, Cheng, Gao, Wang, Gao, and Shan]{xu2024instantmesh}
Jiale Xu, Weihao Cheng, Yiming Gao, Xintao Wang, Shenghua Gao, and Ying Shan.
\newblock Instantmesh: Efficient 3d mesh generation from a single image with sparse-view large reconstruction models.
\newblock \emph{arXiv preprint arXiv:2404.07191}, 2024.

\bibitem[Yang et~al.(2024)Yang, Huang, Guo, Lu, Wu, Lam, Cao, and Liu]{partobjaverse_tiny}
Yunhan Yang, Yukun Huang, Yuan-Chen Guo, Liangjun Lu, Xiaoyang Wu, Edmund~Y. Lam, Yan-Pei Cao, and Xihui Liu.
\newblock Sampart3d: Segment any part in 3d objects, 2024.

\bibitem[Yang et~al.(2025)Yang, Zhou, Guo, Zou, Huang, Liu, Xu, Liang, Cao, and Liu]{yang2025omnipart}
Yunhan Yang, Yufan Zhou, Yuan-Chen Guo, Zi-Xin Zou, Yukun Huang, Ying-Tian Liu, Hao Xu, Ding Liang, Yan-Pei Cao, and Xihui Liu.
\newblock Omnipart: Part-aware 3d generation with semantic decoupling and structural cohesion.
\newblock \emph{arXiv preprint arXiv:2507.06165}, 2025.

\bibitem[Yao et~al.(2025)Yao, Zhang, Yan, Zeng, Zhang, Xu, Yang, Gu, and Yu]{yao2025cast}
Kaixin Yao, Longwen Zhang, Xinhao Yan, Yan Zeng, Qixuan Zhang, Lan Xu, Wei Yang, Jiayuan Gu, and Jingyi Yu.
\newblock Cast: Component-aligned 3d scene reconstruction from an rgb image.
\newblock \emph{ACM Transactions on Graphics (TOG)}, 44\penalty0 (4):\penalty0 1--19, 2025.

\bibitem[Ye et~al.(2025)Ye, Wang, Zhao, Xie, and Zhu]{ye2025shapellmomninativemultimodalllm}
Junliang Ye, Zhengyi Wang, Ruowen Zhao, Shenghao Xie, and Jun Zhu.
\newblock Shapellm-omni: A native multimodal llm for 3d generation and understanding, 2025.

\bibitem[Zhang et~al.(2023)Zhang, Tang, Niessner, and Wonka]{zhang20233dshape2vecset}
Biao Zhang, Jiapeng Tang, Matthias Niessner, and Peter Wonka.
\newblock 3dshape2vecset: A 3d shape representation for neural fields and generative diffusion models.
\newblock \emph{ACM Transactions On Graphics (TOG)}, 42\penalty0 (4):\penalty0 1--16, 2023.

\bibitem[Zhang et~al.(2024)Zhang, Wang, Zhang, Qiu, Pang, Jiang, Yang, Xu, and Yu]{zhang2024clay}
Longwen Zhang, Ziyu Wang, Qixuan Zhang, Qiwei Qiu, Anqi Pang, Haoran Jiang, Wei Yang, Lan Xu, and Jingyi Yu.
\newblock Clay: A controllable large-scale generative model for creating high-quality 3d assets.
\newblock \emph{ACM Transactions on Graphics (TOG)}, 43\penalty0 (4):\penalty0 1--20, 2024.

\bibitem[Zhang et~al.(2025{\natexlab{a}})Zhang, Zhang, Jiang, Bai, Yang, Xu, and Yu]{bang}
Longwen Zhang, Qixuan Zhang, Haoran Jiang, Yinuo Bai, Wei Yang, Lan Xu, and Jingyi Yu.
\newblock Bang: Dividing 3d assets via generative exploded dynamics.
\newblock \emph{ACM Transactions on Graphics}, 44\penalty0 (4):\penalty0 1–21, 2025{\natexlab{a}}.

\bibitem[Zhang et~al.(2018)Zhang, Isola, Efros, Shechtman, and Wang]{lpips}
Richard Zhang, Phillip Isola, Alexei~A. Efros, Eli Shechtman, and Oliver Wang.
\newblock The unreasonable effectiveness of deep features as a perceptual metric, 2018.

\bibitem[Zhang et~al.(2025{\natexlab{b}})Zhang, Li, Zhou, Wu, and Wu]{zhang2025scenelanguagerepresentingscenes}
Yunzhi Zhang, Zizhang Li, Matt Zhou, Shangzhe Wu, and Jiajun Wu.
\newblock The scene language: Representing scenes with programs, words, and embeddings, 2025{\natexlab{b}}.

\bibitem[Zhao et~al.(2025{\natexlab{a}})Zhao, Ye, Wang, Liu, Chen, Wang, and Zhu]{zhao2025deepmeshautoregressiveartistmeshcreation}
Ruowen Zhao, Junliang Ye, Zhengyi Wang, Guangce Liu, Yiwen Chen, Yikai Wang, and Jun Zhu.
\newblock Deepmesh: Auto-regressive artist-mesh creation with reinforcement learning, 2025{\natexlab{a}}.

\bibitem[Zhao et~al.(2025{\natexlab{b}})Zhao, Cao, Xu, Dong, and Shan]{zhao2025assemblerscalable3dassembly}
Wang Zhao, Yan-Pei Cao, Jiale Xu, Yuejiang Dong, and Ying Shan.
\newblock Assembler: Scalable 3d part assembly via anchor point diffusion, 2025{\natexlab{b}}.

\bibitem[Zhao et~al.(2024)Zhao, Liu, Chen, Zeng, Wang, Cheng, Fu, Chen, Yu, and Gao]{zhao2024michelangelo}
Zibo Zhao, Wen Liu, Xin Chen, Xianfang Zeng, Rui Wang, Pei Cheng, Bin Fu, Tao Chen, Gang Yu, and Shenghua Gao.
\newblock Michelangelo: Conditional 3d shape generation based on shape-image-text aligned latent representation.
\newblock \emph{Advances in Neural Information Processing Systems}, 36, 2024.

\bibitem[Zheng et~al.(2025)Zheng, Huang, Chen, and Mao]{zheng2025pro3deditorprogressiveviewsperspective}
Yang Zheng, Mengqi Huang, Nan Chen, and Zhendong Mao.
\newblock Pro3d-editor : A progressive-views perspective for consistent and precise 3d editing, 2025.

\bibitem[Zhu et~al.(2025)Zhu, Zhang, Shao, and Tang]{zhu2025kvedittrainingfreeimageediting}
Tianrui Zhu, Shiyi Zhang, Jiawei Shao, and Yansong Tang.
\newblock Kv-edit: Training-free image editing for precise background preservation, 2025.

\bibitem[Zhuang et~al.(2024)Zhuang, Kang, Cao, Li, Lin, and Shan]{zhuang2024tipeditoraccurate3deditor}
Jingyu Zhuang, Di Kang, Yan-Pei Cao, Guanbin Li, Liang Lin, and Ying Shan.
\newblock Tip-editor: An accurate 3d editor following both text-prompts and image-prompts, 2024.

\end{thebibliography}
}

\clearpage
\setcounter{page}{1}
\maketitlesupplementary

\section{Text-Condition 3D Editing}

\begin{figure}[ht]
    \centering
    \includegraphics[width=\linewidth]{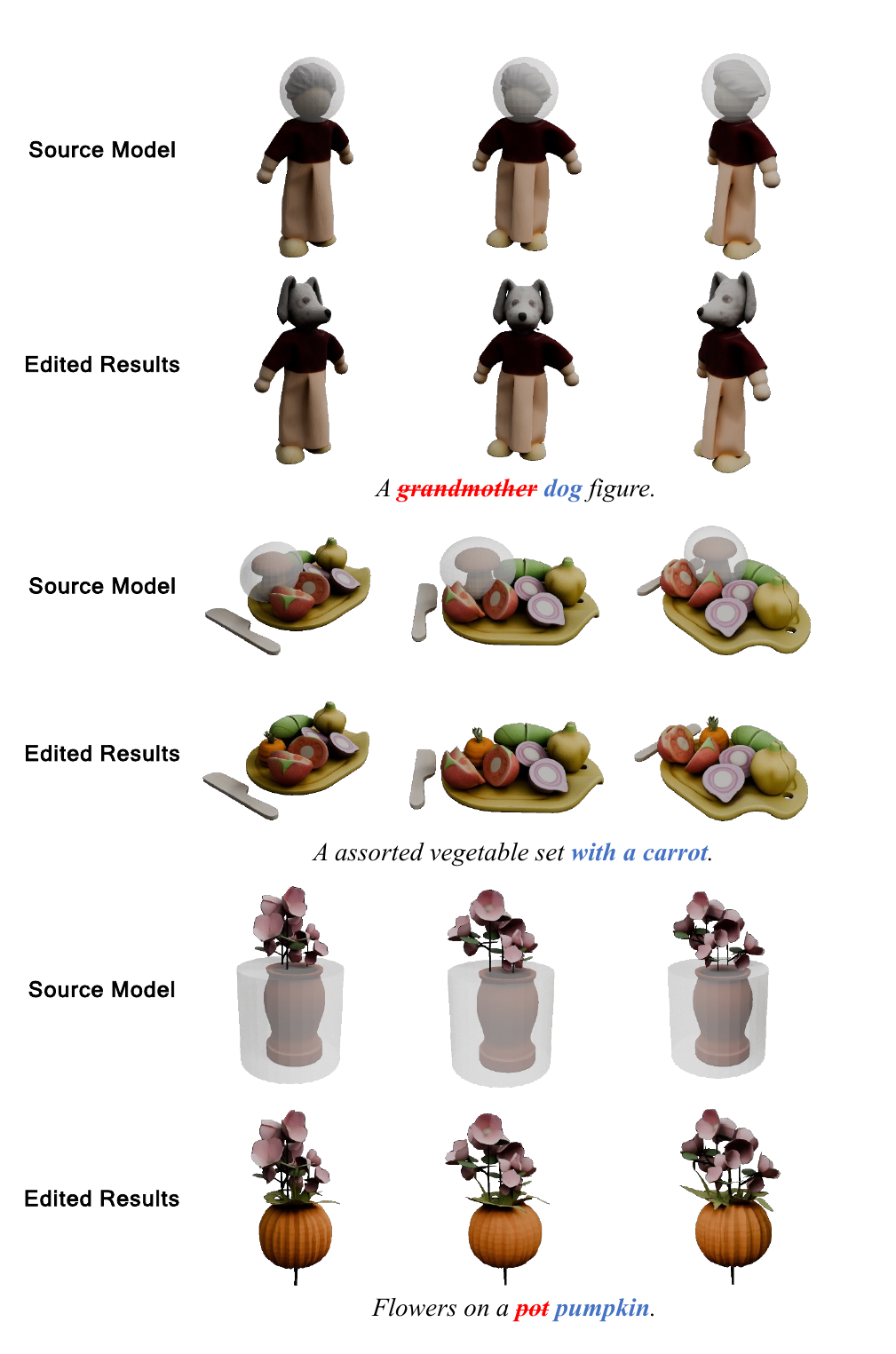}
    \caption{\textbf{Visualization results of text-condition 3D editing.}}
    \label{fig:results_text}
\end{figure}

\begin{figure*}[h]
    \centering
    \includegraphics[width=\textwidth]{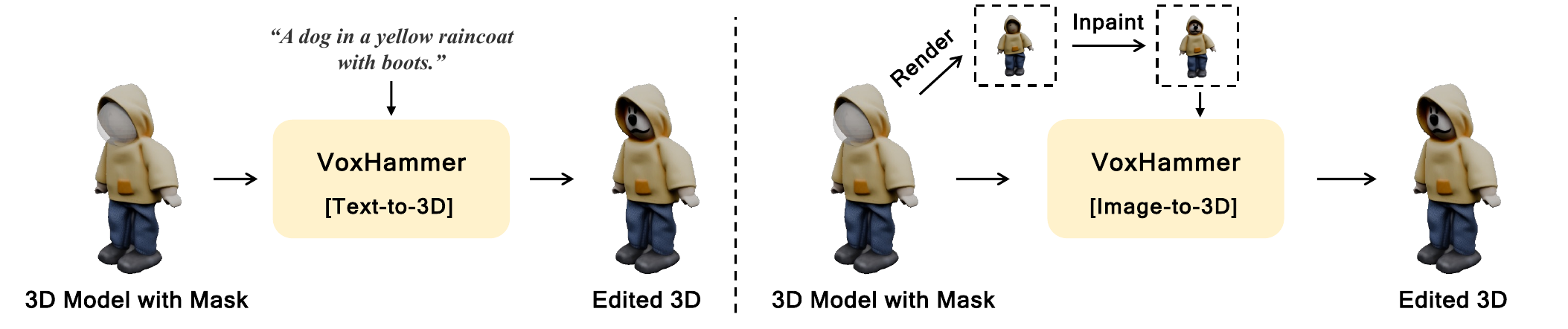}
    \caption{\textbf{Pipeline of text-condition (left) and image-condition (right) 3D editing.}}
    \label{fig:pipeline_full}
\end{figure*}

Benefiting from the versatility of TRELLIS~\cite{trellis}, our framework also supports text-condition 3D editing by injecting textual conditions into the inversion and denoising stages of the base model for masked assets, as illustrated in ~\cref{fig:pipeline_full}.
Leveraging this capability, we evaluate \textit{VoxHammer} on text-condition 3D editing tasks, where it achieves competitive performance in preserving unedited regions and maintaining overall 3D quality, as presented in ~\cref{fig:results_text}. 
However, condition alignment is not always reliable, as the model may deviate from textual instructions. As shown in ~\cref{tab:text_to_3d}, this underscores the need to further enhance the fidelity of text-conditioned guidance.

\section{Explanation of Evaluation Metrics}


In terms of evaluating unedited region preservation, Chamfer Distance assesses the geometry consistency, while masked PSNR, SSIM and LPIPS of rendered multi-view images evaluate the consistency of structures and appearance.
In terms of evaluating editing quality, FID assesses the overall visual similarity between the edited results and the original object.
FVD evaluates the temporal continuity and stability across multi-view images.
In terms of edit controllability, the text-asset alignment score from CLIP-T measures the similarity between the editing results and the editing text, while DINO-I measures the similarity between the editing results and the original object.
Since our task focuses on 3D local editing, DINO-I can reflect the accuracy of the edits to some extent.
Overall, these metrics provide a comprehensive quantitative evaluation of unedited region preservation, overall editing quality, and editing accuracy from different perspectives, collectively reflecting the overall performance of the 3D editing method.

\section{More Results}
\label{sec:more}

\begin{figure*}[h]
    \centering
    \includegraphics[width=0.92\textwidth]{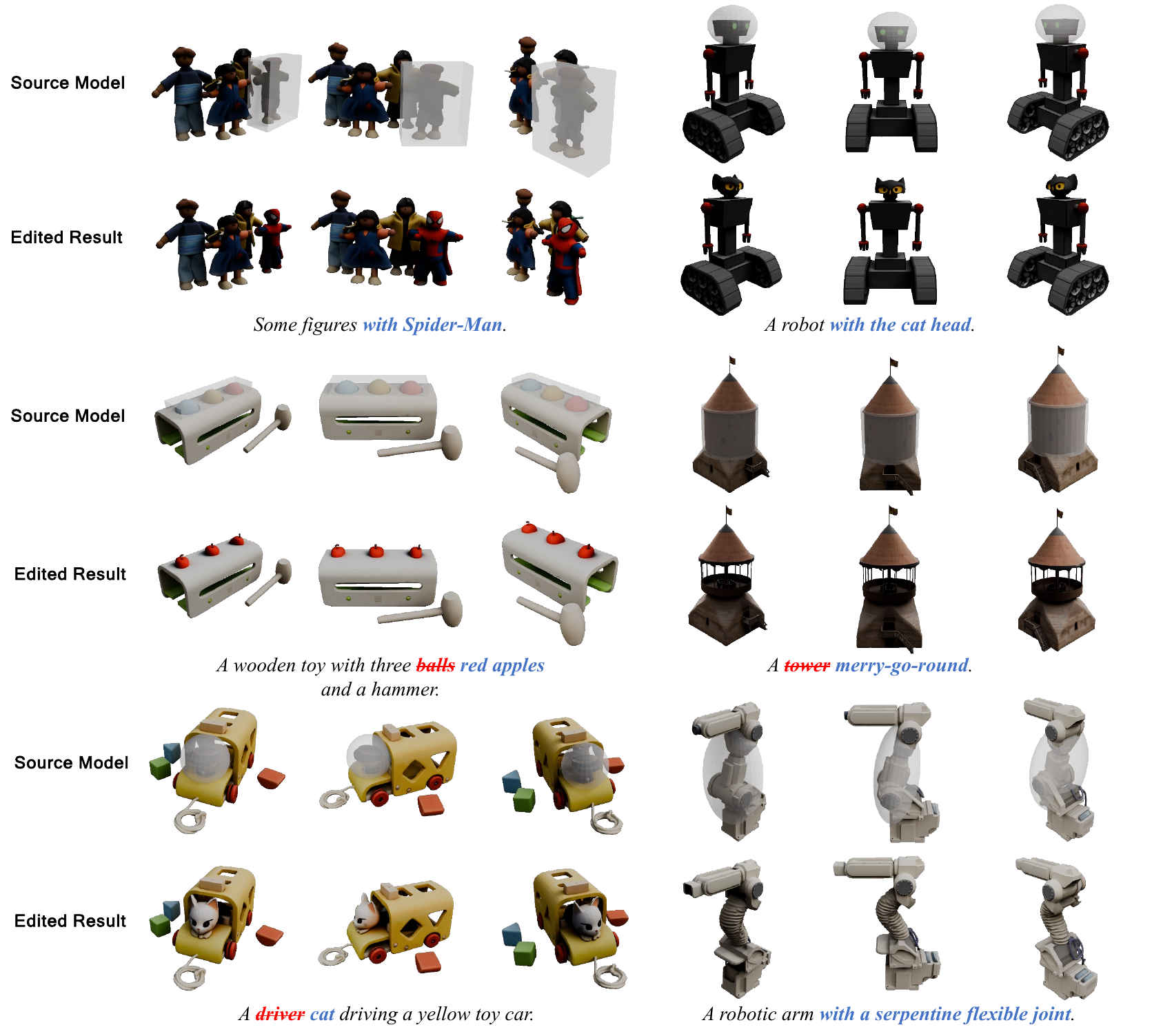}
    \caption{\textbf{More visualization results of image-condition 3D editing.}}
    \label{fig:more_results}
\end{figure*}

More results of image-condition 3D editing are shown in ~\cref{fig:more_results}, which demonstrates the ability to achieve precise and coherent 3D editing.

\section{Limitation}
\label{sec:limitation}

Although \textit{VoxHammer} preserves unedited regions and maintains overall 3D quality, several limitations remain.
First, textual alignment is not yet optimal, partly due to the scarcity of large-scale captioned 3D datasets, making text condition less robust than image-based guidance.
Second, editing fidelity is bounded by the resolution of the TRELLIS~\cite{trellis} backbone, limiting precision for high-resolution assets.
Finally, our pipeline comprises of 3D encoding, inversion, denoising and decoding. Due to the time-consuming rendering phase in the 3D encoding stage ($>$ 1 min), \textit{VoxHammer} takes about 2 minutes to edit one 3D asset, indicating room for efficiency improvements toward interactive use.

\begin{table}[h]
\centering
\small
\caption{\textbf{Runtime comparison across different methods.}}
\label{tab:runtime_comparison}
\begin{tabularx}{0.8\linewidth}{>{\hsize=0.5\hsize}X|>{\hsize=0.5\hsize}X}
\toprule
Method & Runtime \\
\midrule
Vox-E~\cite{sella2023voxetextguidedvoxelediting} & 32 min \\
MVEdit~\cite{mvedit} & 242 s  \\
Tailor3D~\cite{qi2024tailor3dcustomized3dassets} & 83 s  \\
Instant3DiT~\cite{barda2024instant3ditmultiviewinpaintingfast} & \textbf{20 s}  \\
\textbf{Ours} & 133 s \\
\bottomrule
\end{tabularx}
\end{table}

\begin{table*}[h]
\small
\centering
\caption{\textbf{Quantitative comparison on text-condition and image-condition3D  editing.}}
\label{tab:text_to_3d} 
\setlength{\tabcolsep}{3.35mm} %
\resizebox{1.0\linewidth}{!}{
\begin{tabular}{l|cccc|cc|c}
\toprule
\multirow{3}{*}[0.8ex]{Method} & \multicolumn{4}{c|}{Unedited Region Preservation} & \multicolumn{2}{c|}{Overall 3D Quality} &\multicolumn{1}{c}{Condition Alignment} \\
\cmidrule(lr){2-8} & CD. $\downarrow$ & PSNR (M) $\uparrow$ & SSIM (M) $\uparrow$ & LPIPS (M) $\downarrow$ & FID $\downarrow$ & FVD $\downarrow$  & CLIP-T $\uparrow$ \\
\midrule
Text-condition 3D editing  & \textbf{0.010} & 38.61 & 0.992 & \textbf{0.024} & 25.93 & \textbf{150.4} & 0.277 \\
Image-condition 3D editing & 0.012 & \textbf{41.68} & \textbf{0.994} & 0.027 & \textbf{23.05} & 187.8 & \textbf{0.287} \\
\bottomrule
\end{tabular}
}
\end{table*}

\end{document}